\documentclass[11pt,a4paper]{article}
\usepackage[hyperref]{acl2021}
\usepackage{times}
\usepackage{graphicx}
\usepackage{latexsym}
\usepackage{subcaption}
\usepackage{amsmath}
\usepackage{amssymb}
\usepackage{booktabs}
\usepackage{multirow}
\usepackage{enumitem}

\usepackage{microtype}

\newif\ifcomments
\commentstrue
\ifcomments
    \providecommand\ori[1]{[\textcolor{red}{Ori: {#1}}]}
    \providecommand\yuval[1]{[\textcolor{brown}{Yuval: {#1}}]}
    \providecommand\omer[1]{[\textcolor{purple}{Omer: {#1}}]}
    \providecommand\jb[1]{[\textcolor{blue}{Jonathan: {#1}}]}
    \providecommand\amir[1]{[\textcolor{orange}{Amir: {#1}}]}
\else
    \providecommand{\ori}[1]{}
    \providecommand{\yuval}[1]{}
    \providecommand{\omer}[1]{}
    \providecommand{\jb}[1]{}
    \providecommand{\amir}[1]{}
\fi

\aclfinalcopy 

\newcommand{\comment}[1]{}

\title{Few-Shot Question Answering by Pretraining Span Selection}

\author{Ori Ram\thanks{\;\;~~Equal contribution.}$\ \hspace{0.05cm}^{,1}$~~~~Yuval Kirstain$^{*,1}$~~~~Jonathan Berant$^{1,2}$~~~~Amir Globerson$^1$~~~~Omer Levy$^1$\\ \\
Blavatnik School of Computer Science, Tel Aviv University$^1$ \\
Allen Institute for AI$^2$ \\
\texttt{\{ori.ram,yuval.kirstain,joberant,gamir,levyomer\}@cs.tau.ac.il}
}

\date{}

\begin{document}
\maketitle

\begin{abstract}

In several question answering benchmarks, pretrained models have reached human parity through fine-tuning on an order of 100,000 annotated questions and answers.
We explore the more realistic few-shot setting, where only a few hundred training examples are available, and observe that standard models perform poorly, highlighting the discrepancy between current pretraining objectives and question answering. 
We propose a new pretraining scheme tailored for question answering: recurring span selection.
Given a passage with multiple sets of recurring spans, we mask in each set all recurring spans but one, and ask the model to select the correct span in the passage for each masked span.
Masked spans are replaced with a special token, viewed as a question representation, that is later used during fine-tuning to select the answer span.
The resulting model obtains surprisingly good results on multiple benchmarks (e.g., 72.7 F1 on SQuAD with only 128 training examples), while maintaining competitive performance in the high-resource setting.\footnote{Our code, models, and datasets are publicly available: \url{https://github.com/oriram/splinter}.}
\end{abstract}

\section{Introduction}

\begin{figure}[t]
\centering
\hspace*{-20pt}
\includegraphics[width=1.05\columnwidth]{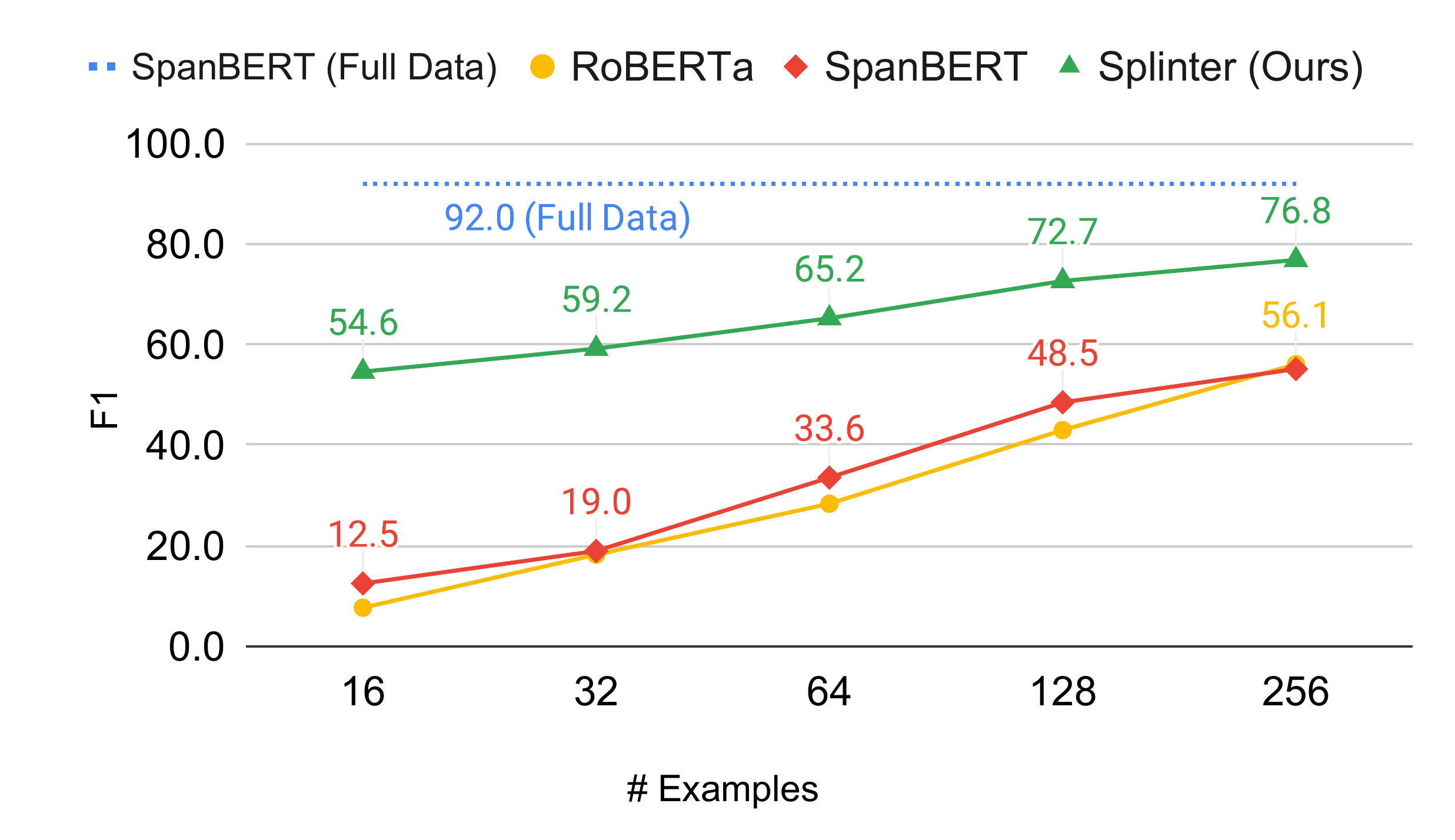}
\caption{Performance of SpanBERT (red) and RoBERTa (yellow) base-size models on SQuAD, given different amounts of training examples. Our model (Splinter, green) dramatically improves performance. SpanBERT-base trained on the \emph{full} training data of SQuAD (blue, dashed) is shown for reference. }
\label{fig:fail}
\end{figure}

\begin{figure*}[t!]
\centering
\includegraphics[width=\textwidth]{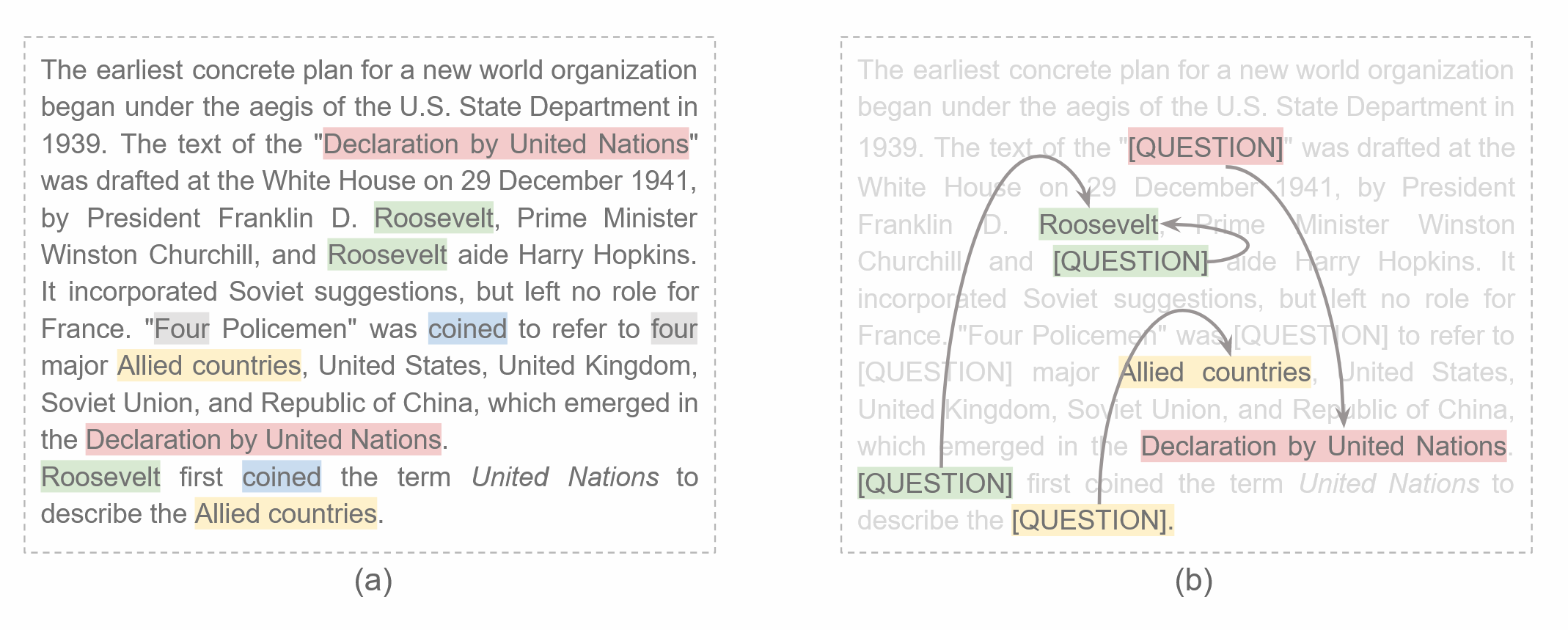}
\caption{An example paragraph before (a) and after (b) masking recurring spans. Each color represents a different cluster of spans. After masking recurring spans (replacing each with a single \texttt{[QUESTION]} token), only one span from each cluster remains unmasked, and is considered the correct answer to the masked spans in the cluster. The pretraining task is to predict the correct answer for each \texttt{[QUESTION]}.}
\label{fig:span_masking}
\end{figure*}

The standard approach to question answering is to pretrain a masked language model on raw text, and then fine-tune it with a span selection layer on top \cite{devlin-etal-2019-bert, joshi-etal-2020-spanbert, liu2019roberta}. 
While this approach is effective, and sometimes exceeds human performance, its success is based on the assumption that large quantities of annotated question answering examples are available. For instance, both SQuAD \cite{rajpurkar-etal-2016-squad,rajpurkar-etal-2018-know} and Natural Questions \cite{kwiatkowski-etal-2019-natural} contain an order of 100,000 question and answer pairs in their training data. 
This assumption quickly becomes unrealistic as we venture outside the lab conditions of 
English Wikipedia, and attempt to crowdsource question-answer pairs in other languages or domains of expertise \cite{tsatsaronis_overview_2015, Kembhavi_2017_CVPR}.
How do question answering models fare in the more practical case, where an in-house annotation effort can only produce a couple hundred training examples?

We investigate the task of few-shot question answering by sampling small training sets from existing question answering benchmarks.
Despite the use of pretrained models, the standard approach yields poor results when fine-tuning on few examples (Figure~\ref{fig:fail}).
For example, RoBERTa-base fine-tuned on 128 question-answer pairs from SQuAD obtains around 40 F1.
This is somewhat expected, since the pretraining objective is quite different from the fine-tuning task. While masked language modeling requires mainly \textit{local} context around the masked token, question answering needs to align the question with the \textit{global} context of the passage.
To bridge this gap, we propose (1) a novel self-supervised method for pretraining span selection models, and (2) a question answering layer that aligns a representation of the question with the text.

We introduce \textit{Splinter} (\textbf{sp}an-\textbf{l}evel po\textbf{inter}), a pretrained model for few-shot question answering.
The challenge in defining such a self-supervised task is how to create question-answer pairs from unlabeled data. Our key observation is that one can leverage \textit{recurring spans}: n-grams, such as named entities, which tend to occur multiple times in a given passage (e.g., \textit{``Roosevelt''} in Figure~\ref{fig:span_masking}). We emulate question answering by masking all but one instance of each recurring span with a special \texttt{[QUESTION]} token, and asking the model to select the correct span for each such token.

\begin{figure*}[t]
\centering
\includegraphics[width=\textwidth]{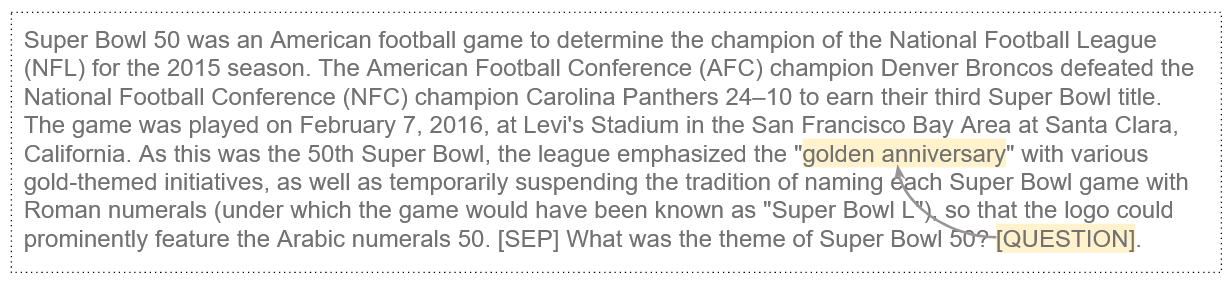}
\caption{An example of our fine-tuning setup, taken from the development set of SQuAD. The question, followed by the \texttt{[QUESTION]} token, is concatenated to the context. The \texttt{[QUESTION]} token's representation is then used to select the answer span.}
\label{fig:question}
\end{figure*}

To select an answer span for each \texttt{[QUESTION]} token \textit{in parallel}, we introduce a question-aware span selection (QASS) layer, which uses 
the \texttt{[QUESTION]} token's representation to select the answer span. The QASS layer seamlessly integrates with fine-tuning on real question-answer pairs. We simply append the \texttt{[QUESTION]} token to the input question, and use the QASS layer to select the answer span (Figure~\ref{fig:question}). This is unlike existing models for span selection, which do not include an explicit question representation. The compatibility between pretraining and fine-tuning makes Splinter an effective few-shot learner.

Splinter exhibits surprisingly high performance given only a few training examples throughout a variety of benchmarks from the MRQA 2019 shared task \cite{fisch-etal-2019-mrqa}.
For example, Splinter-base achieves 72.7 F1 on SQuAD with only 128 examples, outperforming all baselines by a very wide margin.
An ablation study shows that the pretraining method and the QASS layer itself (even without pretraining) both contribute to improved performance. 
Analysis indicates that Splinter's representations change significantly less during fine-tuning compared to the baselines, suggesting that our pretraining is more adequate for question answering.
Overall, our results highlight the importance of designing objectives and architectures in the few-shot setting, where an appropriate inductive bias can lead to dramatic performance improvements.

\section{Background}
\label{sec:background}

Extractive question answering is a common task in NLP, where the goal is to select a contiguous span $a$ from a given text $T$ that answers a question $Q$.
This format was popularized by SQuAD \cite{rajpurkar-etal-2016-squad}, and has since been adopted by several datasets in various domains \cite{trischler-etal-2017-newsqa,Kembhavi_2017_CVPR} and languages \cite{lewis-etal-2020-mlqa,clark-etal-2020-tydi},
with some extensions allowing for unanswerable questions \cite{levy-etal-2017-zero, rajpurkar-etal-2018-know} or multiple answer spans \cite{dua-etal-2019-drop, dasigi-etal-2019-quoref}.
In this work, we follow the assumptions in the recent MRQA 2019 shared task \cite{fisch-etal-2019-mrqa} and focus on questions whose answer is a single span.

The standard approach uses a pretrained encoder, such as BERT \cite{devlin-etal-2019-bert}, and adds two parameter vectors $\mathbf{s}, \mathbf{e}$ to the pretrained model in order to detect the start position $s$ and end position $e$ of the answer span $a$, respectively.
The input text $T$ and question $Q$ are concatenated and fed into the encoder, producing a contextualized token representation $\mathbf{x}_i$ for each token in the sequence.
To predict the start position of the answer span, a probability distribution is induced over the entire sequence by computing the inner product of a learned vector $\mathbf{s}$ with every token representation (the end position is computed similarly using a vector $\mathbf{e}$):
\begin{align*}
P( s = i \mid T, Q) &= \frac{\exp(\mathbf{x}_i ^\top \mathbf{s})}{\sum_j \exp(\mathbf{x}_j ^\top \mathbf{s})}, \\
P( e = i \mid T, Q) &= \frac{\exp(\mathbf{x}_i ^\top \mathbf{e})}{\sum_j \exp(\mathbf{x}_j ^\top \mathbf{e})}.
\end{align*}
The parameters $\mathbf{s}, \mathbf{e}$ are trained during fine-tuning, using the cross-entropy loss with the start and end positions of the gold answer span.

This approach assumes that each token representation $\mathbf{x}_i$ is contextualized with respect to the question.
However, the masked language modeling objective does not necessarily encourage this form of long-range contextualization in the pretrained model, since many of the masked tokens can be resolved from local cues.
Fine-tuning the attention patterns of pretrained masked language models may thus entail an extensive learning effort, difficult to achieve with only a handful of training examples.
We overcome this issue by (1) pretraining directly for span selection, and (2) explicitly representing the question with a single vector, used to detect the answer in the input text.
\section{Splinter}
\label{sec:model}

We formulate a new task for pretraining question answering from unlabeled text: \textit{recurring span selection}.
We replace spans that appear multiple times in the given text with a special \texttt{[QUESTION]} token, except for one occurrence, which acts as the ``answer'' span for each (masked) cloze-style ``question''.
The prediction layer is a modification of the standard span selection layer, which replaces the static start and end parameter vectors, $\mathbf{s}$ and $\mathbf{e}$, with dynamically-computed boundary detectors based on the contextualized representation of each \texttt{[QUESTION]} token.
We reuse this architecture when fine-tuning on question-answer pairs by adding a \texttt{[QUESTION]} token at the end of the actual question, thus aligning the pretraining objective with the fine-tuning task.
We refer to our pretrained model as \textit{Splinter}.

\subsection{Pretraining: Recurring Span Selection}
\label{subsec:pretraining}

Given an input text $T$, we find all \textit{recurring spans}: arbitrary n-grams that appear more than once in the same text.
For each set of identical recurring spans $R$, we select a single occurrence as the \textit{answer} $a$ and replace all other occurrences with a single \texttt{[QUESTION]} token.\footnote{In practice, only some sets of recurring spans are processed; see \textit{Cluster Selection} below.}
The goal of recurring span selection is to predict the correct answer $a$ for a given \texttt{[QUESTION]} token $q \in R \setminus \{a\}$, each $q$ thus acting as an independent \textit{cloze-style question}.

Figure~\ref{fig:span_masking} illustrates this process.
In the given passage, the span \textit{``Roosevelt''} appears three times.
Two of its instances (the second and third) are replaced with \texttt{[QUESTION]}, while one instance (the first) becomes the answer, and remains intact.  
After masking, the sequence is passed through a transformer encoder, producing contextualized token representations.
The model is then tasked with predicting the start and end positions of the answer given each \texttt{[QUESTION]} token representation.
In Figure~\ref{fig:span_masking}b, we observe four instances of this prediction task: two for the \textit{``Roosevelt''} cluster, one for the \textit{``Allied countries''} cluster, and one for \textit{``Declaration by United Nations''}.

Taking advantage of recurring \emph{words} in a passage (restricted to nouns or named entities) was proposed in past work as a signal for coreference \cite{kocijan-etal-2019-wikicrem,ye-etal-2020-coreferential}. We further discuss this connection in Section~\ref{sec:related}.



\paragraph{Span Filtering}
To focus pretraining on semantically meaningful spans, we use the following definition for ``spans'', which filters out recurring spans that are likely to be uninformative:
(1) spans must begin and end at word boundaries,
(2) we consider only maximal recurring spans,
(3) spans containing only stop words are ignored,
(4) spans are limited to a maximum of 10 tokens.
These simple heuristic filters do not require a model, as opposed to masking schemes in related work \cite{glass-etal-2020-span, ye-etal-2020-coreferential, Guu2020REALMRL}, which require part-of-speech taggers, constituency parsers, or named entity recognizers.

\paragraph{Cluster Selection}
We mask a random subset of recurring span clusters in each text, leaving some recurring spans untouched.
Specifically, we replace up to $30$ spans with \texttt{[QUESTION]} from each input passage.\footnote{In some cases, the last cluster may have more than one unmasked span.}
This number was chosen to resemble the 15\% token-masking ratio of \citet{joshi-etal-2020-spanbert}. Note that in our case, the number of masked tokens is greater than the number of questions.

\subsection{Model: Question-Aware Span Selection}
\label{subsec:architecture}

Our approach converts texts into a set of questions that need to be answered 
simultaneously. The standard approach for extractive question answering \cite{devlin-etal-2019-bert} is inapplicable, because it uses fixed start and end vectors. Since we have multiple questions, we replace the standard parameter vectors $\mathbf{s}, \mathbf{e}$ with \textit{dynamic} start and end vectors $\mathbf{s}_q , \mathbf{e}_q$, computed from each \texttt{[QUESTION]} token $q$:
\begin{align*}
\mathbf{s}_q = \mathbf{S} \mathbf{x}_q \qquad \mathbf{e}_q = \mathbf{E} \mathbf{x}_q
\end{align*}
Here, $\mathbf{S} , \mathbf{E}$ are parameter matrices, which extract ad hoc start and end position detectors $\mathbf{s}_q , \mathbf{e}_q$ from the given \texttt{[QUESTION]} token's representation $\mathbf{x}_q$. The rest of our model follows the standard span selection model by computing the start and end position probability distributions.
The model can also be viewed as two bilinear functions of the question representation $\mathbf{x}_q$ with each token in the sequence $\mathbf{x}_i$, similar to \citet{dozat2017deep}:
\begin{align*}
P( s = i \mid T, q) &= \frac{\exp(\mathbf{x}_i^\top \mathbf{S} \mathbf{x}_q)}{\sum_j \exp(\mathbf{x}_j^\top \mathbf{S} \mathbf{x}_q)} \\
P( e = i \mid T, q) &= \frac{\exp(\mathbf{x}_i^\top \mathbf{E} \mathbf{x}_q)}{\sum_j \exp(\mathbf{x}_j^\top \mathbf{E} \mathbf{x}_q)}
\end{align*}
Finally, we use the answer's gold start and end points $(s_a, e_a)$ to compute the cross-entropy loss:
\begin{align*}
- \log P(s = s_a \mid T, q) - \log P(e = e_a \mid T, q)
\end{align*}
We refer to this architecture as the question-aware span selection (QASS) layer.

\subsection{Fine-Tuning}
\label{subsec:fine_tuning}

After pretraining, we assume access to labeled examples, where each training instance is a text $T$, a question $Q$, and an answer $a$ that is a span in $T$.
To make this setting similar to pretraining, we simply append a \texttt{[QUESTION]} token to the input sequence, immediately after the question $Q$ (see Figure~\ref{fig:question}). Selecting the answer span then proceeds exactly as during pretraining. Indeed, the advantage of our approach is that in both pretraining and fine-tuning, the \texttt{[QUESTION]} token representation captures information about the question that is then used to select the span from context.

\section{A Few-Shot QA Benchmark}
\label{sec:fewshot}

To evaluate how pretrained models work when only a small amount of labeled data is available for fine-tuning, we simulate various low-data scenarios by sampling subsets of training examples from larger datasets.
We use a subset of the MRQA 2019 shared task \cite{fisch-etal-2019-mrqa}, which contains extractive question answering datasets in a unified format, where the answer is a single span in the given text passage.

Split I of the MRQA shared task contains 6 large question answering datasets: SQuAD \cite{rajpurkar-etal-2016-squad}, NewsQA \cite{trischler-etal-2017-newsqa}, TriviaQA \cite{joshi-etal-2017-triviaqa}, SearchQA \cite{dunn2017searchqa}, HotpotQA \cite{yang-etal-2018-hotpotqa}, and Natural Questions \cite{kwiatkowski-etal-2019-natural}.
For each dataset, we sample smaller training datasets from the original training set with sizes changing on a logarithmic scale, from 16 to 1,024 examples.
To reduce variance, for each training set size, we sample 5 training sets using different random seeds and report average performance across training sets.
We also experiment with fine-tuning the models on the full training sets.
Since Split I of the MRQA shared task does not contain test sets, we evaluate using the official development sets as our test sets.

We also select two datasets from Split II of the MRQA shared task that were annotated by domain experts: BioASQ \cite{tsatsaronis_overview_2015} and TextbookQA \cite{Kembhavi_2017_CVPR}.
Each of these datasets only has a development set that is publicly available in MRQA, containing about 1,500 examples.
For each dataset, we sample 400 examples for evaluation (test set), and follow the same protocol we used for large datasets to sample training sets of 16 to 1,024 examples from the remaining data.

To maintain the few-shot setting, every dataset in our benchmark has well-defined training and test sets.
To tune hyperparameters, one needs to extract validation data from each training set.
For simplicity, we do not perform hyperparameter tuning or model selection (see Section 5), and thus use all of the available few-shot data for training.

\begin{figure*}[t]
\centering
\hspace*{-20pt}
\includegraphics[width=1.05\columnwidth]{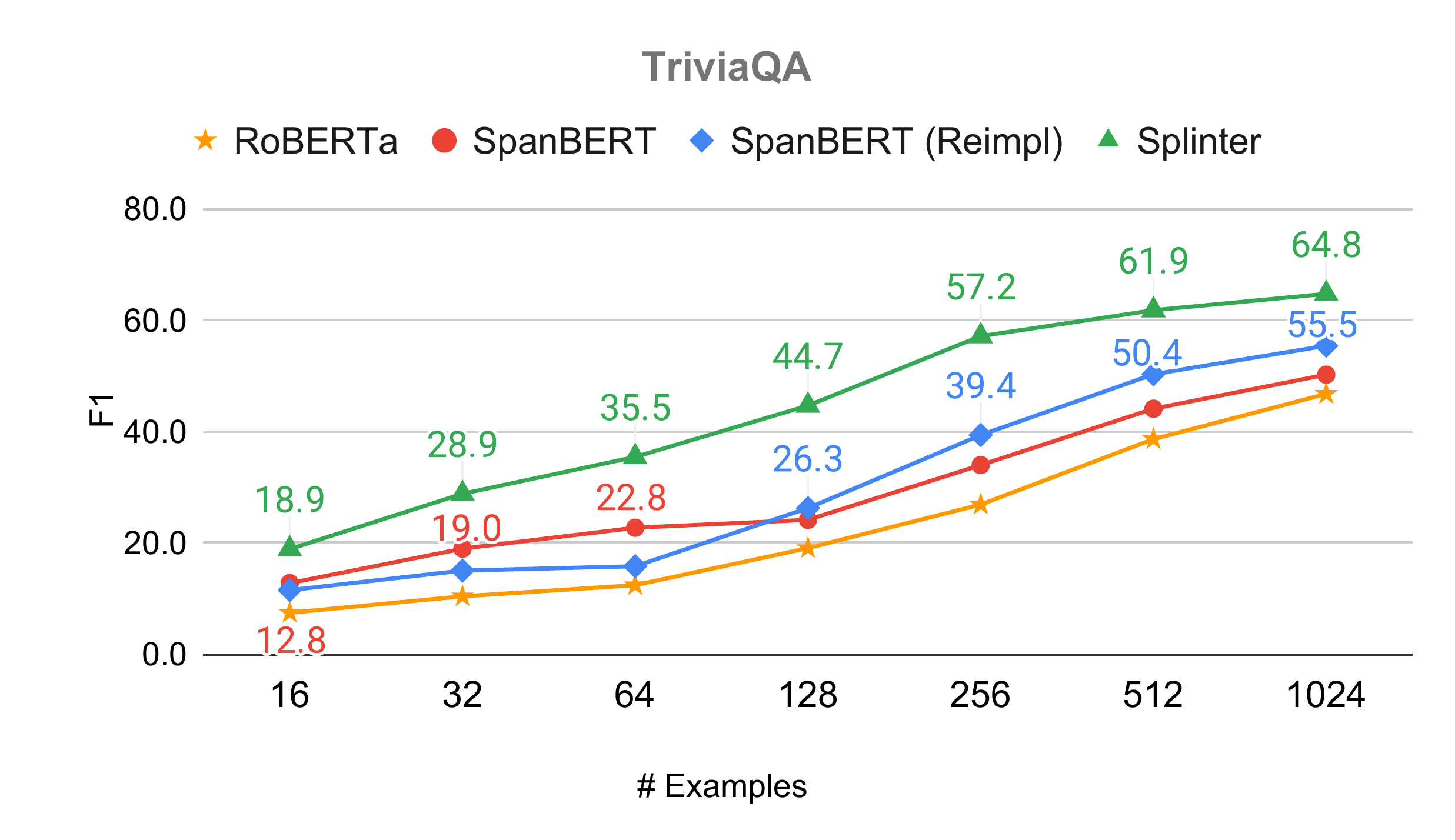} 
\includegraphics[width=1.05\columnwidth]{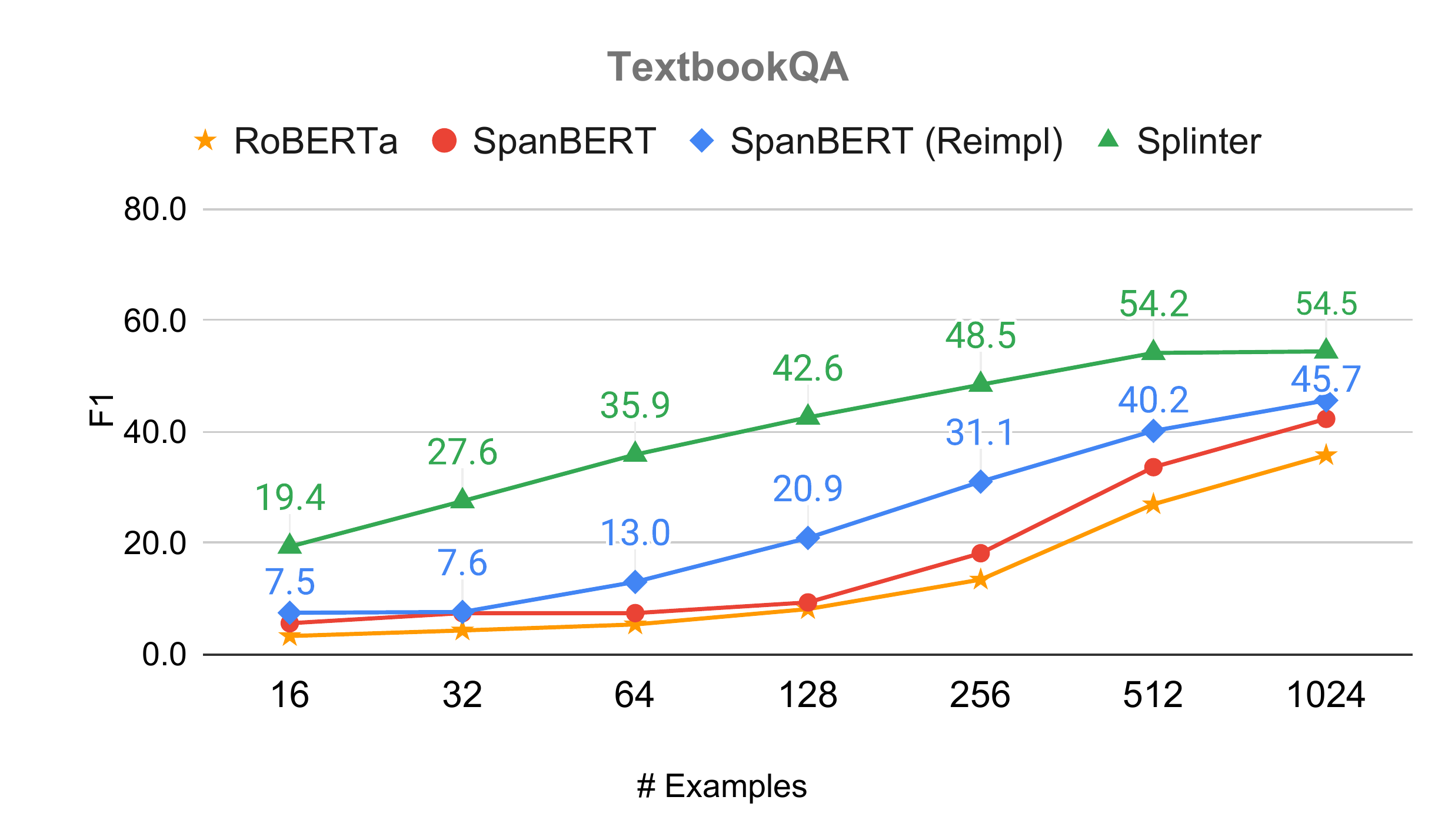} 
\caption{Performance (F1) of Splinter-base (green), compared to all baselines as a function of the number of training examples on two 
datasets. Each point reflects the average performance across 5 randomly-sampled training sets of the same size.}
\label{fig:fewshot}
\end{figure*}

\section{Experimental Setup}
\label{sec:setup}

We describe our experimental setup in detail, including all models and baselines. 

\subsection{Baselines}
\label{sec:baselines}

Splinter-base shares the same architecture (transformer encoder \cite{vaswaniNIPS2017_7181}), vocabulary (cased wordpieces), and number of parameters (110M) with SpanBERT-base \cite{joshi-etal-2020-spanbert}.
In all experiments, we compare Splinter-base to three baselines of the same capacity:

\paragraph{RoBERTa}
\cite{liu2019roberta} A highly-tuned and optimized version of BERT, which is known to perform well on a wide range of natural language understanding tasks.

\paragraph{SpanBERT}
\cite{joshi-etal-2020-spanbert} A BERT-style model that
focuses on span representations. SpanBERT is trained by masking contiguous spans of tokens and optimizing two objectives: (a) masked language modeling, which predicts each masked token from its own vector representation; (b) the span boundary objective, which predicts each masked token from the representations of the unmasked tokens at the start and end of the masked span.

\paragraph{SpanBERT (Reimpl)}
Our reimplementation of SpanBERT, using exactly the same code, data, and hyperparameters as Splinter.
This baseline aims to control for implementation differences and measures the effect of replacing masked language modeling with recurring span selection.
Also, this version does not use the span boundary objective, as \citet{joshi-etal-2020-spanbert} reported no significant improvements from using it in question answering.

\subsection{Pretraining Implementation}

We train Splinter-base using Adam \cite{kingma2017adam} for 2.4M training steps with batches of 256 sequences of length 512.\footnote{We used this setting to approximate SpanBERT's hyperparameter setting in terms of epochs. That said, SpanBERT-base was trained for a quarter of the steps (600k steps) using four times as many examples per batch (1024 sequences). See Section~\ref{sec:baselines} for additional baselines that control for this difference.}
The learning rate is warmed up for 10k steps to a maximum value of $10^{-4}$, after which it decays linearly. 
As in previous work, we use a dropout rate of 0.1 across all layers.

We follow \citet{devlin-etal-2019-bert} and train on English Wikipedia (preprocessed by WikiExtractor as in  \citet{Wikiextractor2015}) and the Toronto BookCorpus \cite{Zhu_2015_bookcorpus}.
We base our implementation on the official TensorFlow implementation of BERT, and train on a single eight-core v3 TPU (v3-8) on the Google Cloud Platform.

\subsection{Fine-Tuning Implementation}

For fine-tuning, we use the hyperparameters from the default configuration of the HuggingFace Transformers package \cite{wolf-etal-2020-transformers}.\footnote{We did rudimentary tuning on the number of steps only, using a held-out portion of the SQuAD training set, since our training sets can be too small for the default values (e.g., running 10 epochs on 16 examples results in 20 update steps). }
Specifically, we train all models using Adam \cite{kingma2017adam} with bias-corrected moment estimates for few-shot learning \cite{zhang2020revisiting}.
When fine-tuning on 1024 examples or less,  we train for either 10 epochs or 200 steps (whichever is larger). For full-size datasets, we train for 2 epochs.
We set the batch size to 12 and use a maximal learning rate of $3 \cdot 10^{-5}$, which warms up in the first 10\% of the steps, and then decays linearly. 

An interesting question is how to fine-tune the QASS layer parameters (i.e., the $\mathbf{S}$ and $\mathbf{E}$ matrices in Section~\ref{subsec:architecture}). 
In our implementation, we chose to discard the pretrained values and fine-tune from a random initialization, due to the possible discrepancy between span statistics in pretraining and fine-tuning datasets.
However, we report results on fine-tuning without resetting the QASS parameters as an ablation study (Section~\ref{subsec:ablation}).

\begin{table*}[t]
\small
\centering
\begin{tabular}{@{}lcccccccc@{}}
\toprule
\textbf{Model} & \small\textbf{SQuAD} & \small\textbf{TriviaQA} & \small\textbf{NQ} & \small\textbf{NewsQA} & \small\textbf{SearchQA} & \small\textbf{HotpotQA} & \small\textbf{BioASQ} & \small\textbf{TextbookQA} \\
\midrule
\small\textit{16 Examples}  & & & & & & & &  \\
\midrule
RoBERTa & ~~7.7 & ~~7.5 & 17.3 & ~~1.4 & ~~6.9 & 10.5 & 16.7 & ~~3.3 \\
SpanBERT & 12.5 & 12.8 & 19.7 & ~~6.0 & 13.0 & 12.6 & 22.0 & ~~5.6 \\
SpanBERT (Reimpl) & 18.2 & 11.6 & 19.6 & ~~7.6 & 13.3 & 12.5 & 15.9 & ~~7.5   \\
\textbf{Splinter} & \textbf{54.6} & \textbf{18.9} & \textbf{27.4} & \textbf{20.8} & \textbf{26.3} & \textbf{24.0} & \textbf{28.2} & \textbf{19.4}\\
\midrule
\small\textit{128 Examples}  & & & & & & & & \\
\midrule
 RoBERTa & 43.0 & 19.1 & 30.1 & 16.7 & 27.8 & 27.3 & 46.1 & ~~8.2 \\
SpanBERT & 48.5 & 24.2 & 32.2 & 17.4 & 34.3 & 35.1 & 55.3 & ~~9.4 \\
SpanBERT (Reimpl) & 55.8 & 26.3 & 36.0 & 29.5 & 26.3 & 36.6 & 52.2 & 20.9 \\
\textbf{Splinter} & \textbf{72.7} & \textbf{44.7} &\textbf{46.3} & \textbf{43.5} & \textbf{47.2} & \textbf{54.7} & \textbf{63.2} & \textbf{42.6} \\
\midrule
\small\textit{1024 Examples}  & & & & & & & & \\
\midrule
 RoBERTa & 73.8&46.8&54.2&47.5&54.3&61.8&84.1&35.8\\
SpanBERT & 77.8&50.3&57.5&49.3&60.1&67.4&89.3&42.3\\
SpanBERT (Reimpl) & 77.8&55.5&59.5&52.2&58.9&64.6&89.0&45.7\\
\textbf{Splinter} & \textbf{82.8} &\textbf{64.8} &\textbf{65.5} &\textbf{57.3} &\textbf{67.3}&\textbf{70.3}&\textbf{91.0}&\textbf{54.5}\\
\midrule
\small\textit{Full Dataset}  & & & & & & & & \\
\midrule
 RoBERTa & 90.3 &	74.0 &	79.6 & 69.8 & 81.5 &	78.7 & - & -  \\
SpanBERT & 92.0&\textbf{77.2}&80.6&\textbf{71.3} & 80.1 &	79.6 & - & - \\
SpanBERT (Reimpl) & 92.0&75.8&80.5&71.1&81.4&79.7& - & - \\
\textbf{Splinter} & \textbf{92.2}&76.5&\textbf{81.0}&\textbf{71.3}&\textbf{83.0}&\textbf{80.7}& - & - \\
\bottomrule
\end{tabular}
\caption{Performance (F1) across all datasets when the number of training examples is 16, 128, and 1024. We also show performance when training on the full-sized large datasets (MRQA version). All models have the same capacity to BERT-base (110M parameters). NQ stands for Natural Questions.}
\label{tab:fewshot}
\end{table*}

\section{Results}

Our experiments show that Splinter dramatically improves performance in the challenging few-shot setting, unlocking the ability to train question answering models with only hundreds of examples.
When trained on large datasets with an order of 100,000 examples, Splinter is competitive with (and often better than) the baselines.
Ablation studies demonstrate the contributions of 
both recurring span selection pretraining and the QASS layer.

\subsection{Few-Shot Learning}
\label{subsec:few-shot-results}

Figure~\ref{fig:fewshot} shows the F1 score \cite{rajpurkar-etal-2016-squad} of Splinter-base, plotted against all baselines for two datasets, TriviaQA and TextbookQA, as a function of the number of training examples (see Figure~\ref{fig:additional_results} in the appendix for the remaining datasets).
In addition, Table~\ref{tab:fewshot} shows the performance of individual models when given 16, 128, and 1024 training examples across all datasets (see Table~\ref{tab:fewshot_full} in the appendix for additional performance and standard deviation statistics).
It is evident that Splinter outperforms all baselines by large margins.

Let us examine the results on SQuAD, for example.
Given 16 training examples, Splinter obtains 54.6 F1, significantly higher than the best baseline's 18.2 F1.
When the number of training examples is 128, Splinter achieves 72.7 F1, outperforming the baselines by 17 points (our reimplementation of SpanBERT) to 30 (RoBERTa).
When considering 1024 examples, there is a 5-point margin between Splinter (82.8 F1) and SpanBERT (77.8 F1).
The same trend is seen in the other datasets, whether they are in-domain sampled from larger datasets (e.g. TriviaQA) or not;
in TextbookQA, for instance, we observe absolute gaps of 9 to 23 F1 between Splinter and the next-best baseline.

\subsection{High-Resource Regime}

Table~\ref{tab:fewshot} also shows the performance when fine-tuning on the entire training set, when an order of 100,000 examples are available.
Even though Splinter was designed for few-shot question answering, it reaches the best result in five out of six datasets.
This result suggests that when the target task is extractive question answering, it is better to pretrain with our recurring span selection task than with masked langauge modeling, regardless of the number of annotated training examples.

\begin{figure*}[t]
\hspace*{-20pt}
\centering
\includegraphics[width=1.05\columnwidth]{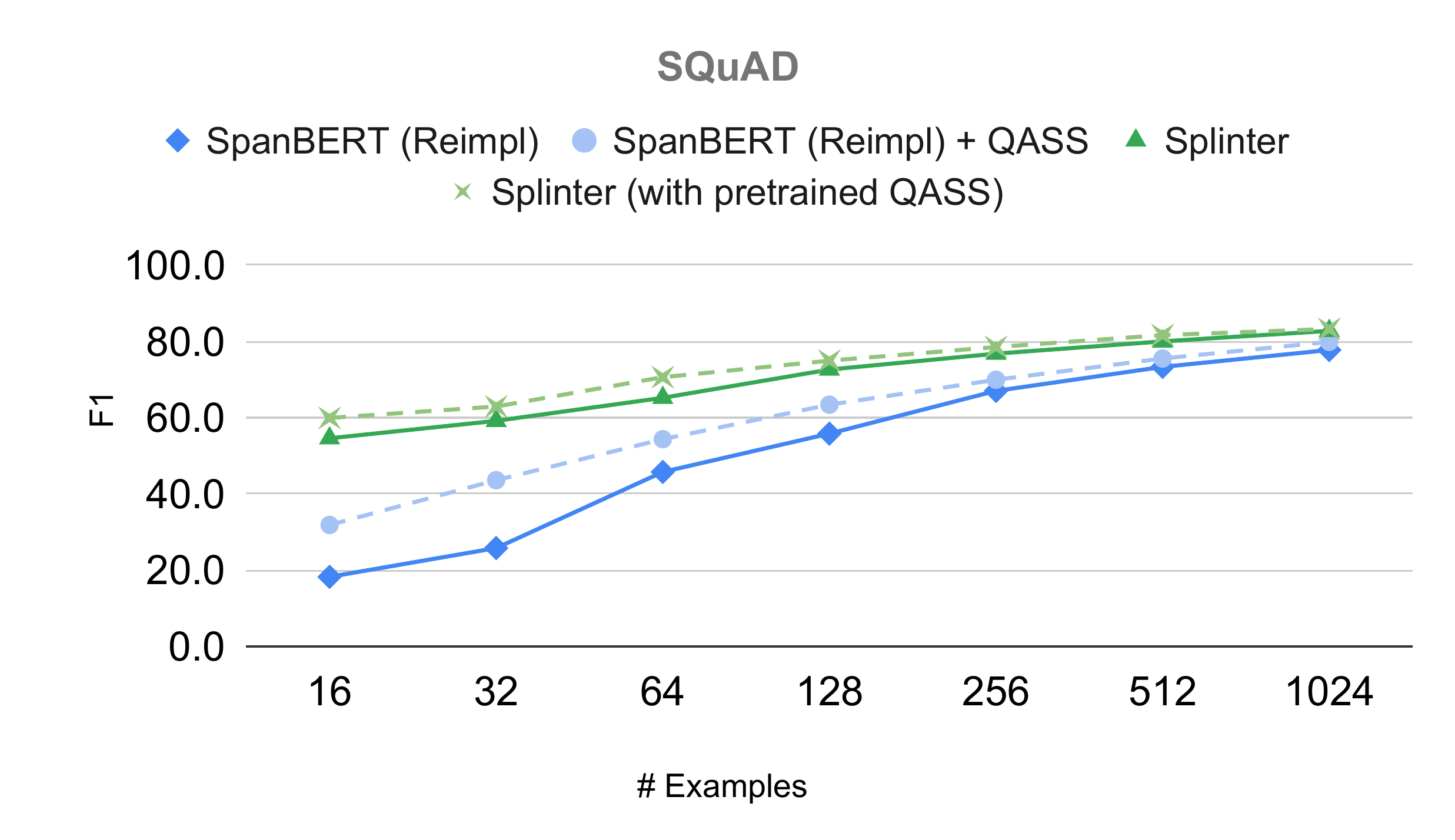}
\includegraphics[width=1.05\columnwidth]{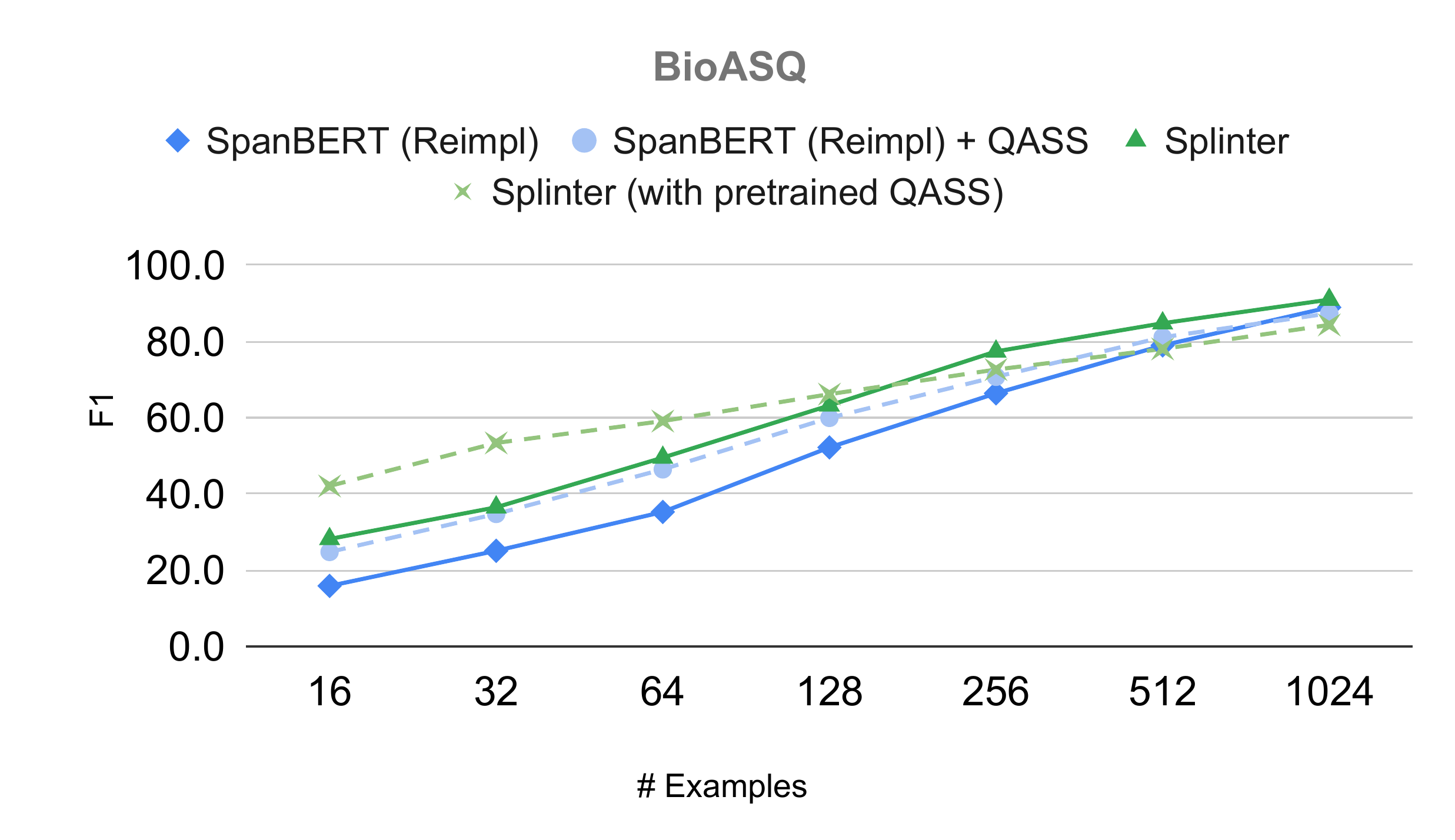} 
\caption{Ablation studies on SQuAD and BioASQ datasets. We examine the role of the QASS layer by fine-tuning it on top of our reimplementation of SpanBERT. In addition, we test whether it is beneficial to keep the pretrained parameters of the QASS layer when fine-tuning Splinter.}
\label{fig:ablation}
\end{figure*}

\subsection{Ablation Study}
\label{subsec:ablation}

We perform an ablation study to better understand the independent contributions of the pretraining scheme and the QASS layer.
We first ablate the effect of pretraining on recurring span selection by applying the QASS layer to pretrained masked language models.
We then test whether the QASS layer's pretrained parameters can be reused in Splinter during fine-tuning without reinitializion.

\paragraph{Independent Contribution of the QASS Layer}
While the QASS layer is motivated by our pretraining scheme, it can also be used without pretraining.
We apply a randomly-initialized QASS layer to our implementation of SpanBERT, and fine-tune it in the few-shot setting.
Figure~\ref{fig:ablation} shows the results of this ablation study for two datasets (see Figure~\ref{fig:ablation_appendix} in the appendix for more datasets).
We observe that replacing the static span selection layer with QASS can significantly improve performance on few-shot question answering.
Having said that, most of Splinter's improvements in the extremely low data regime do stem from combining the QASS layer with our pretraining scheme, and this combination still outperforms all other variants as the amount of data grows.


\paragraph{QASS Reinitialization}
Between pretraining and fine-tuning, we randomly reinitialize the parameters of the QASS layer.
We now test the effect of fine-tuning with the QASS layer's pretrained parameters;
intuitively, the more similar the pretraining data is to the task, the better the pretrained layer will perform.
Figure~\ref{fig:ablation} shows that the advantage of reusing the pretrained QASS is data-dependent, and can result in both performance gains (e.g. extremely low data in SQuAD) and stagnation (e.g. BioASQ with 256 examples or more).
Other datasets exhibit similar trends (see appendix).
We identify three conditions that determine whether keeping the pretrained head is preferable: (1) when the number of training examples is extremely low, (2) when the target domain is similar to that used at pretraining (e.g. Wikipedia), and (3) when the questions are relatively simple (e.g. SQuAD versus HotpotQA).
The latter two conditions pertain to the compatibility between pretraining and fine-tuning tasks; the information learned in the QASS layer is useful as long as the input and output distribution of the task are close to those seen at pretraining time.

\begin{table}[t!]
\small
\centering
\begin{tabular}{@{}lc@{}}
\toprule
Model & Representation Similarity \\
\midrule
RoBERTa & 0.29 \\
SpanBERT  & 0.23 \\
SpanBERT (Reimpl) & 0.19 \\
\textbf{Splinter}  & \textbf{0.89} \\

\bottomrule
\end{tabular}
\caption{Cosine similarity of the representations produced by the transformer encoder before and after fine-tuning on 128 SQuAD examples.}
\label{table:analysis}
\end{table}





\subsection{Analysis}

The recurring span selection objective was designed to emulate extractive question answering using unlabeled text.
How similar is it to the actual target task?
To answer this question, we measure how much each pretrained model's functionality has changed after fine-tuning on 128 examples of SQuAD.
For the purpose of this analysis, we measure change in functionality by examining the vector representation of each token as produced by the transformer encoder; specifically, we measure the cosine similarity between the vector produced by the pretrained model and the one produced by the fine-tuned model, given exactly the same input.
We average these similarities across every token of 200 examples from SQuAD's test set.

Table~\ref{table:analysis} shows that Splinter's outputs are very similar before and after fine-tuning (0.89 average cosine similarity), while the other models' representations seem to change drastically.
This suggests that fine-tuning with even 128 question-answering examples makes significant modifications to the functionality of pretrained masked language models.
Splinter's pretraining, on the other hand, is much more similar to the fine-tuning task, resulting in much more modest changes to the produced vector representations.

\section{Related Work}\label{sec:related}

The remarkable results of GPT-3 \cite{brown2020language} have inspired a renewed interest in few-shot learning.
While some work focuses on classification tasks \cite{schick2020exploiting, gao2020making}, our work investigates few-shot learning in the context of extractive question answering.

One approach to this problem is to create synthetic text-question-answer examples.
Both \citet{lewis-etal-2019-unsupervised} and \citet{glass-etal-2020-span} use the traditional NLP pipeline to select noun phrases and named entities in Wikipedia paragraphs as potential answers, which are then masked from the context to create pseudo-questions.
\citet{lewis-etal-2019-unsupervised} use methods from unsupervised machine translation to translate the pseudo-questions into real ones, while \citet{glass-etal-2020-span} keep the pseudo-questions but use information retrieval to find new text passages that can answer them.
Both works assume access to language- and domain-specific NLP tools such as part-of-speech taggers, syntactic parsers, and named-entity recognizers, which might not always be available.
Our work deviates from this approach by exploiting the natural phenomenon of \textit{recurring spans} in order to generate multiple question-answer pairs per text passage, without assuming any language- or domain-specific models or resources are available beyond plain text.

Similar ideas to recurring span selection were used for creating synthetic coreference resolution examples \cite{kocijan-etal-2019-wikicrem, varkel-globerson-2020-pre}, which mask single words that occur multiple times in the same context.
CorefBERT \cite{ye-etal-2020-coreferential} combines this approach with a copy mechanism for predicting the masked word during pretraining, alongside the masked language modeling objective.
Unlike our approach, which was designed to align well with span selection, CorefBERT masks only \textit{single-word nouns} (rather than arbitrary spans) and replaces each token in the word with a separate mask token (rather than a single mask for the entire multi-token word).
Therefore, it does not emulate extractive question answering.
We did not add CorefBERT as a baseline since the performance of both CorefBERT-base and CorefBERT-large was lower than SpanBERT-base's performance on the full-data MRQA benchmark, and pretraining CorefBERT from scratch was beyond our available computational resources.

\section{Conclusion}

We explore the few-shot setting of extractive question answering, and demonstrate that existing methods, based on fine-tuning large pretrained language models, fail in this setup. 
We propose a new pretraining scheme and architecture for span selection that lead to dramatic improvements, reaching surprisingly good results even when only an order of a hundred examples are available.
Our work shows that choices that are often deemed unimportant when enough data is available, again become crucial in the few-shot setting, opening the door to new methods that take advantage of prior knowledge on the downstream task during model development.


\section*{Acknowledgements}
This project was funded by the
European Research Council (ERC) under the European
Unions Horizon 2020 research and innovation programme
(grant ERC HOLI 819080), the Blavatnik Fund, the Alon Scholarship, the Yandex Initiative for Machine Learning and Intel Corporation. We thank Google’s TPU Research Cloud (TRC) for their support in providing TPUs for this research.

\bibliographystyle{acl_natbib}
\bibliography{anthology,acl2021}

\appendix

\section{Additional Results}
\label{appendix:results}

\begin{figure*}[t!]
\centering
\includegraphics[width=\columnwidth]{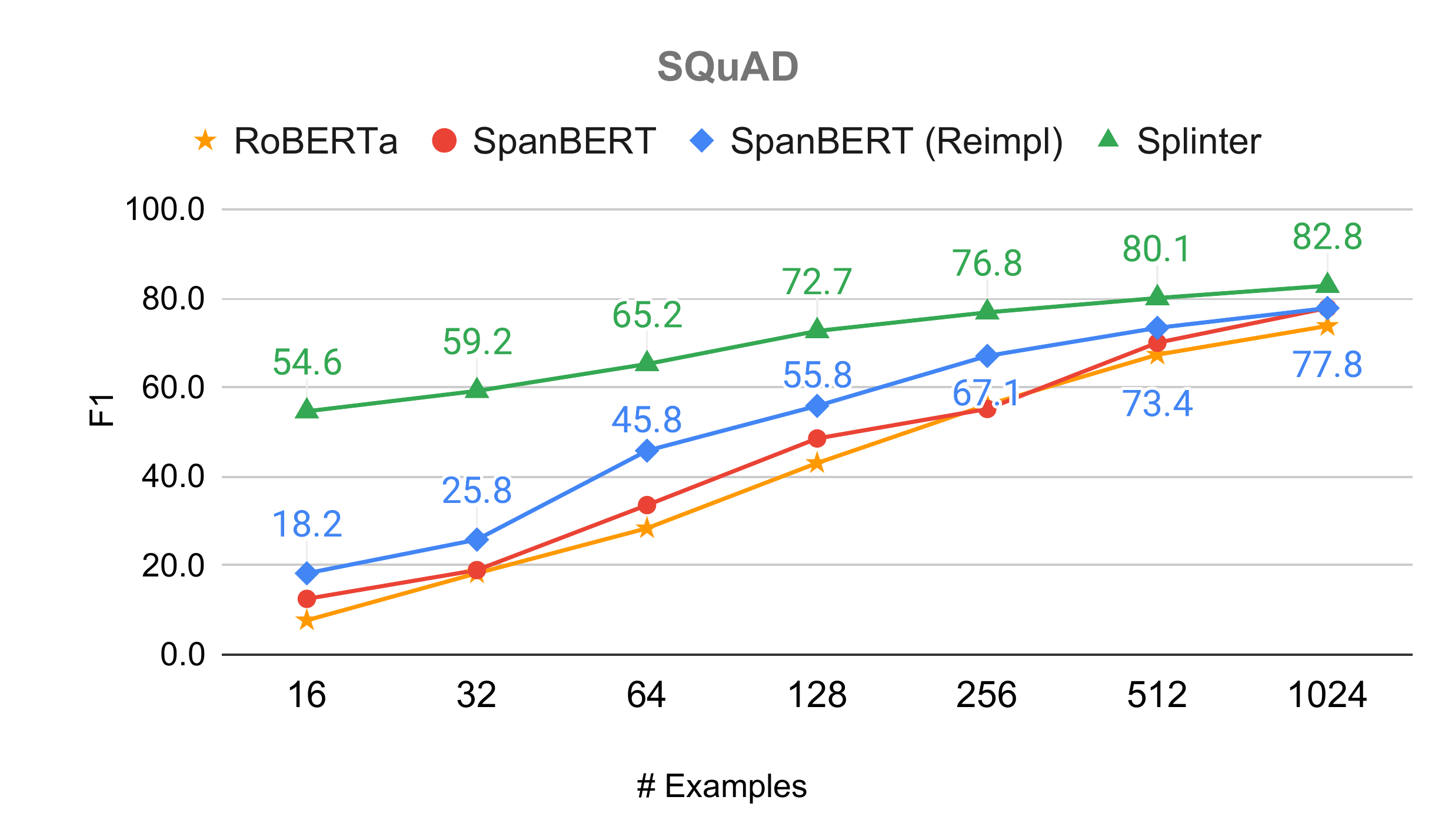}
\includegraphics[width=\columnwidth]{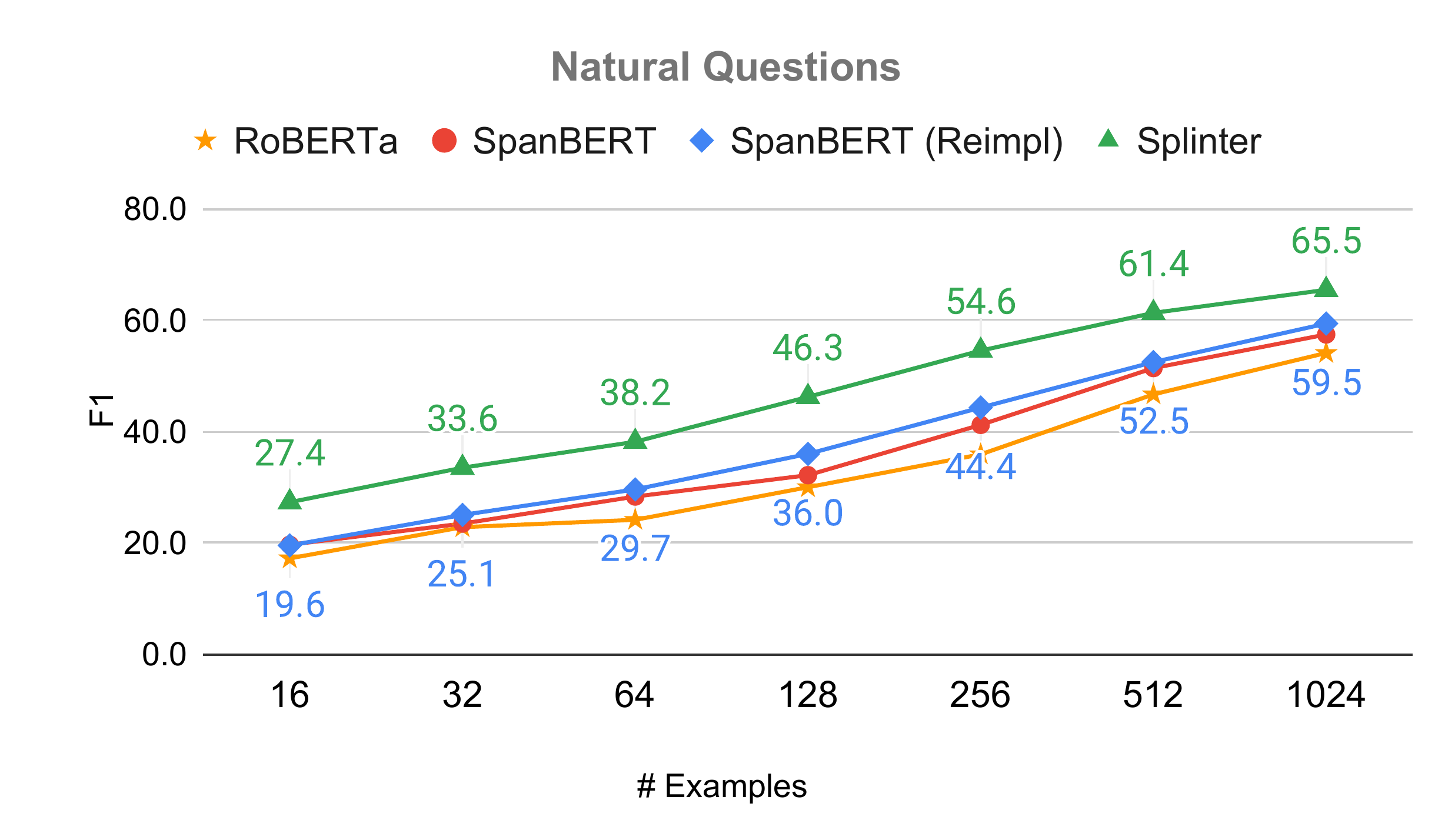}
\includegraphics[width=\columnwidth]{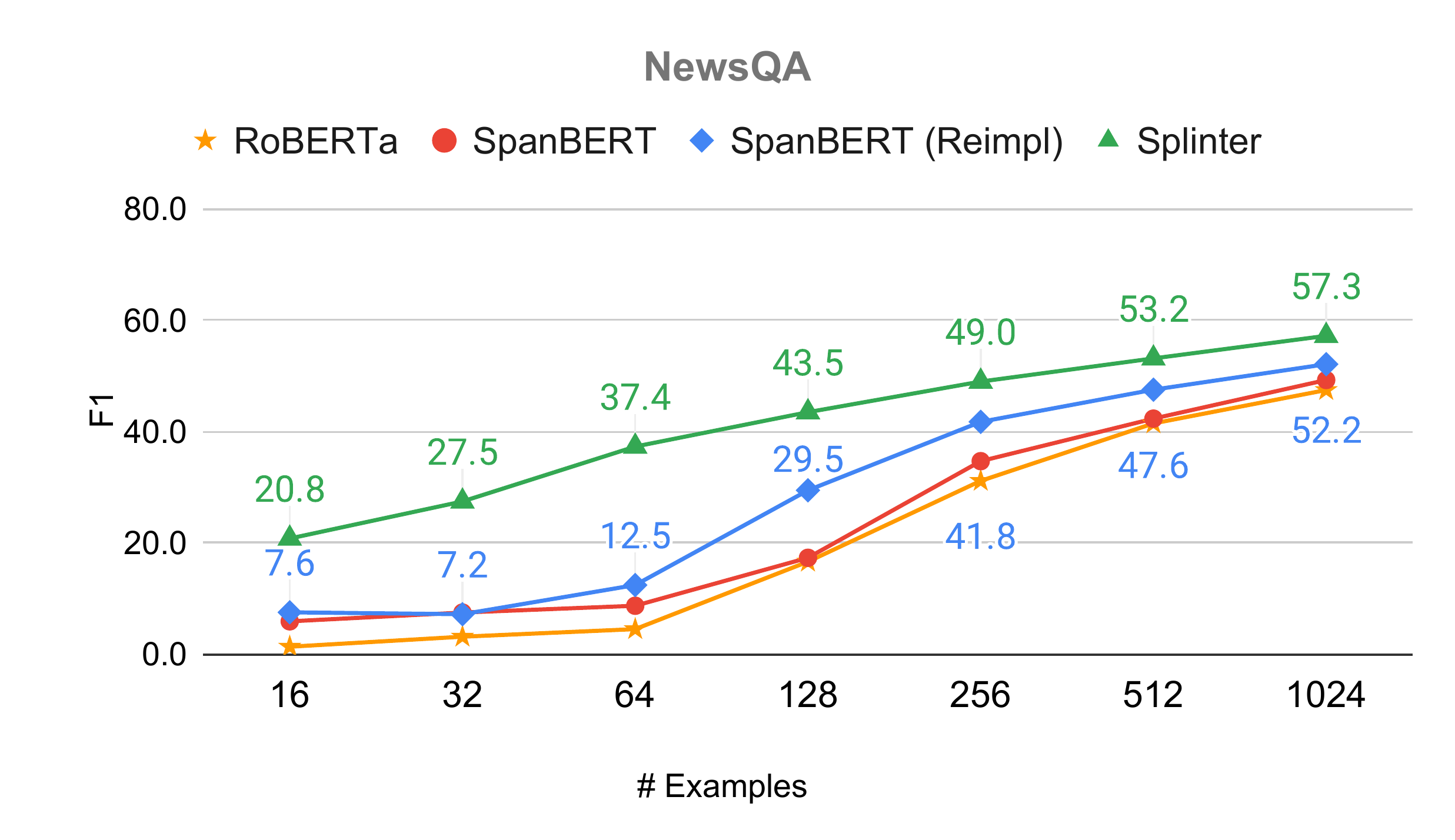}
\includegraphics[width=\columnwidth]{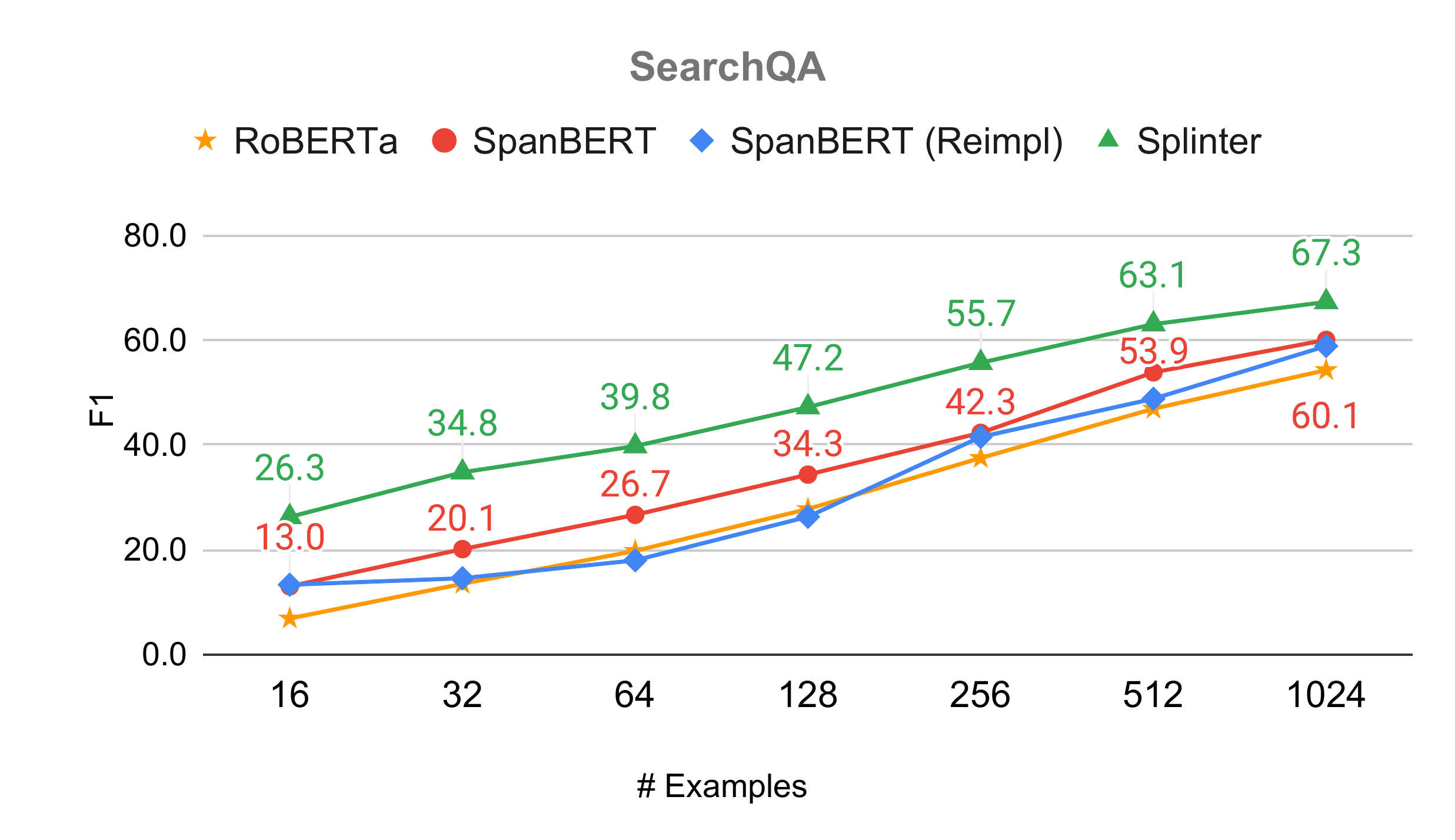} 
\includegraphics[width=\columnwidth]{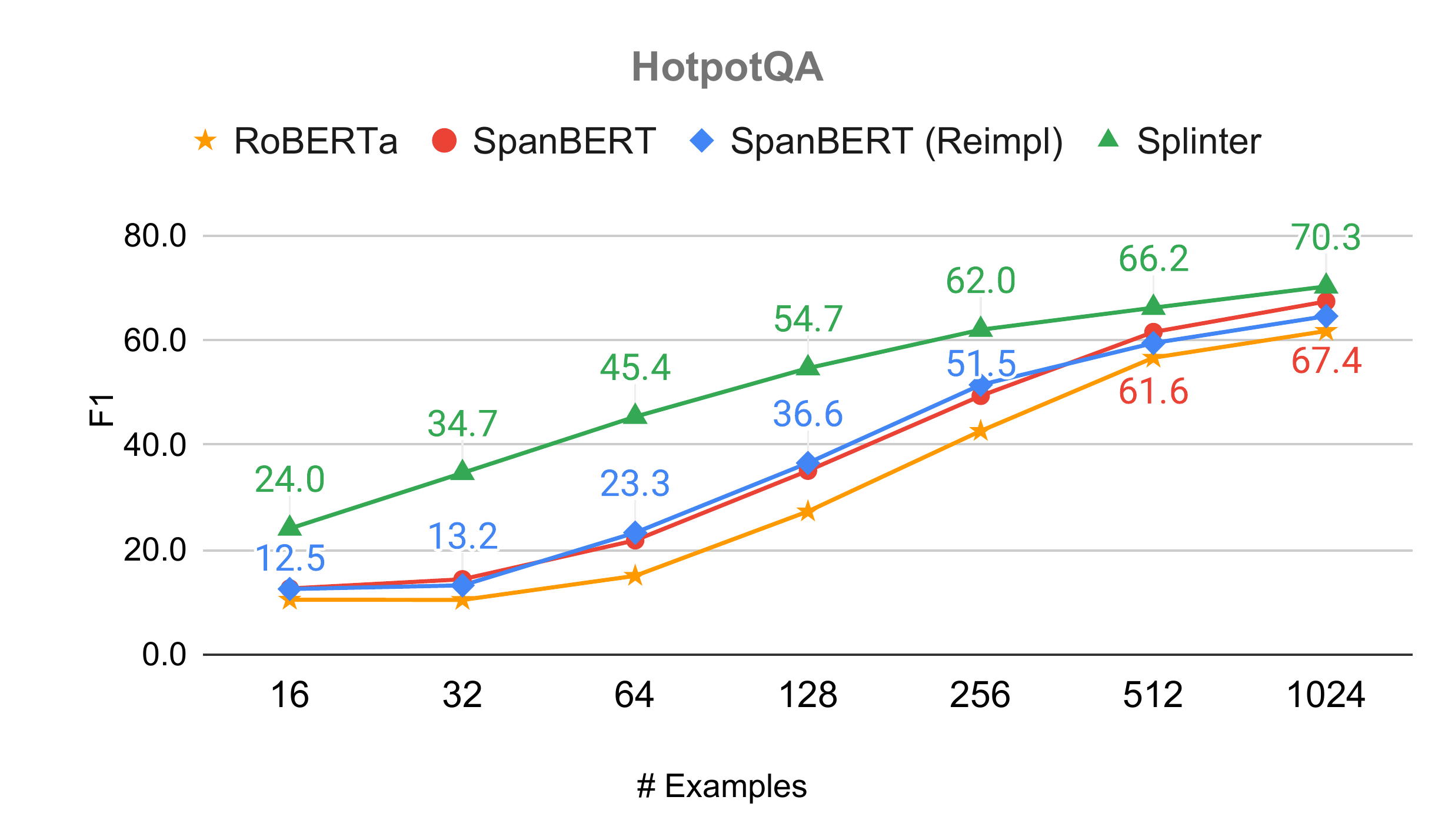} 
\includegraphics[width=\columnwidth]{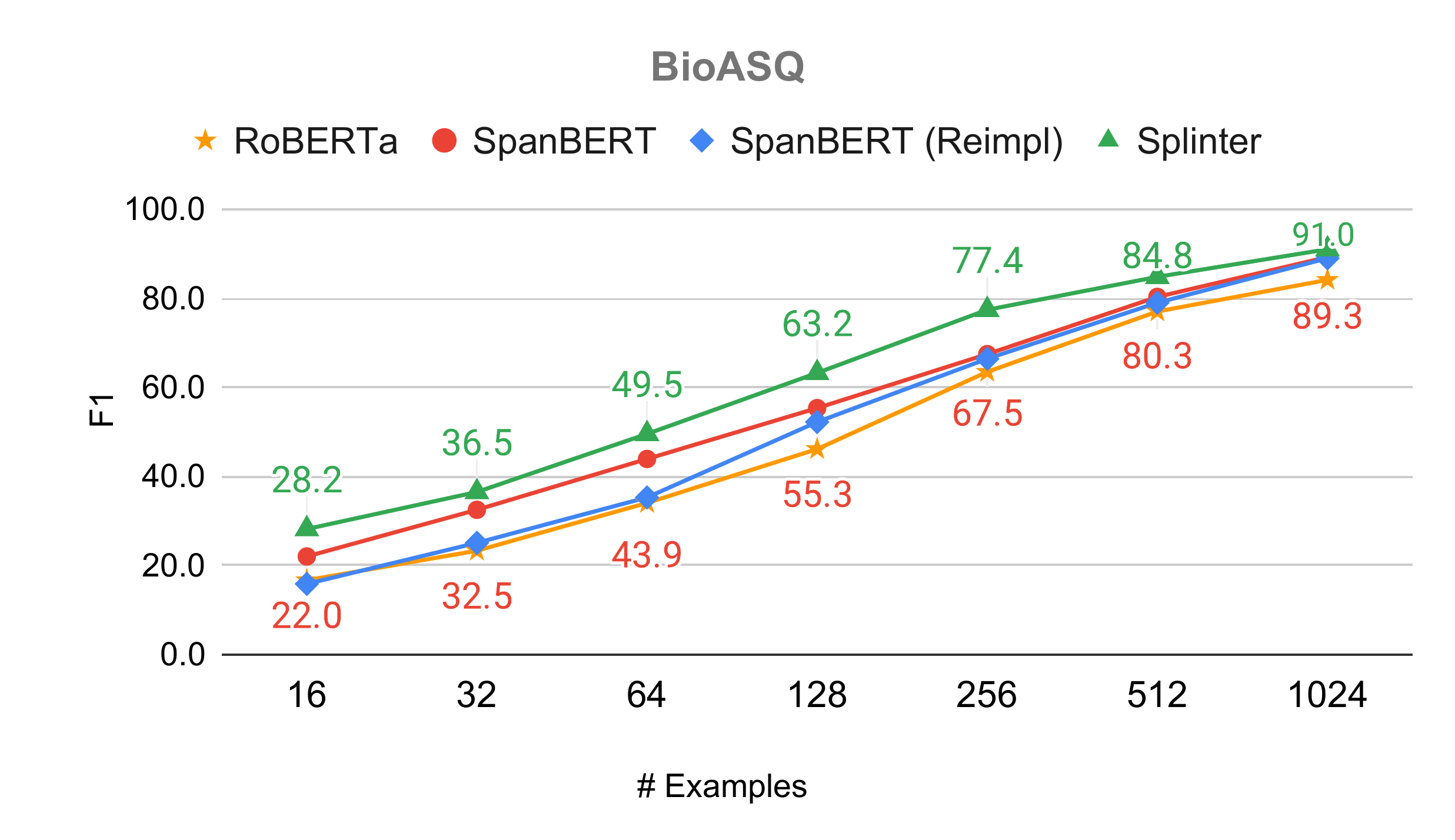} 
\caption{Results complementary to Table 1. Performance (F1) of Splinter-base (green line, triangular points), compared to all baselines as a function of the number of training examples on 4 datasets. 
Each point reflects the average performance across 5 randomly-sampled training sets of the same size.}
\label{fig:additional_results}
\end{figure*}

\paragraph{Few-Shot Results} Figure~\ref{fig:additional_results} shows the results on the six few-shot question answering datasets not included in Figure~\ref{fig:fewshot}. In addition, we give the full raw results (including standard deviation) in Table~\ref{tab:fewshot_full}.

\begin{figure*}[t]
\centering
\includegraphics[width=\columnwidth]{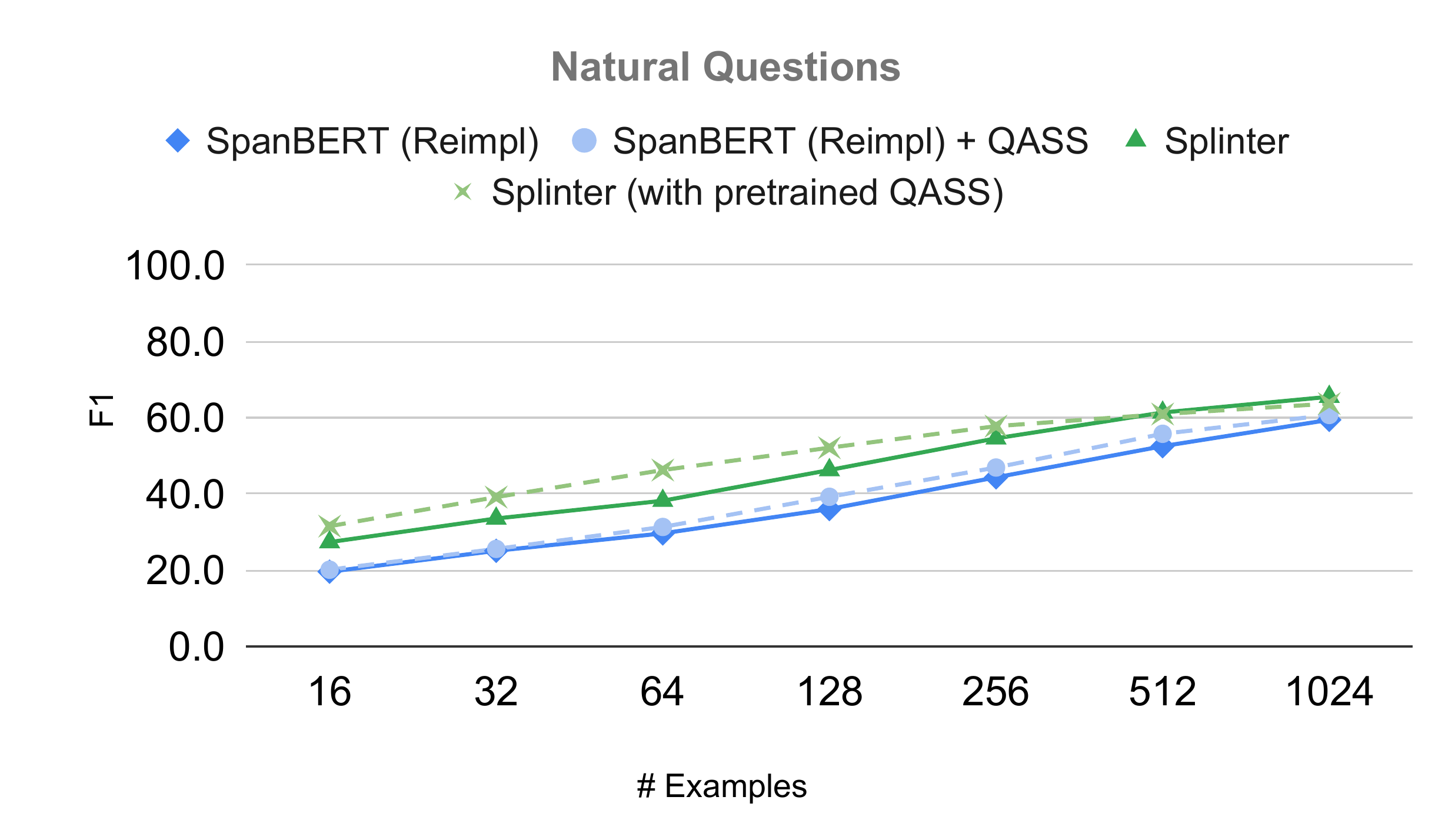}
\includegraphics[width=\columnwidth]{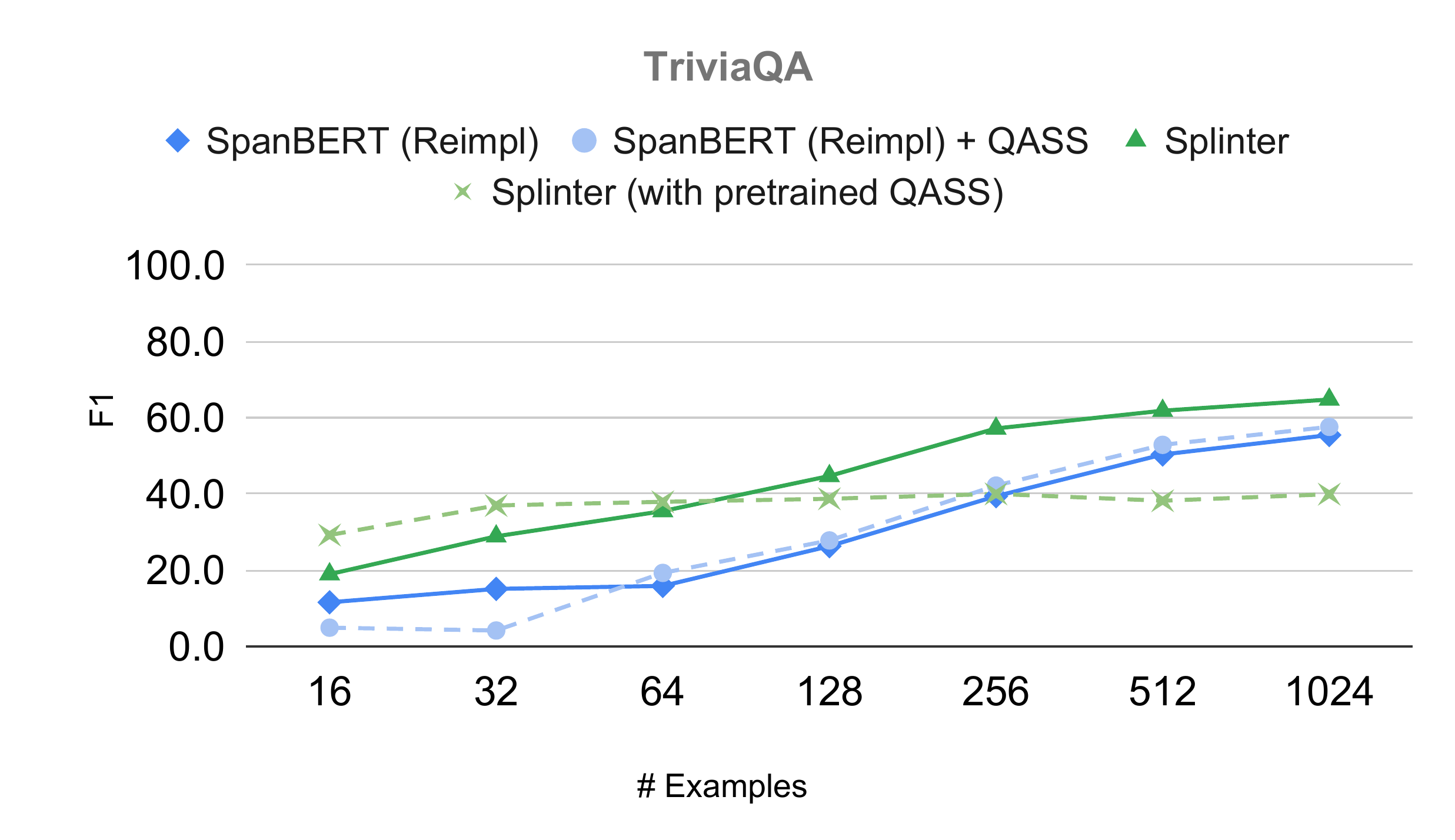}
\includegraphics[width=\columnwidth]{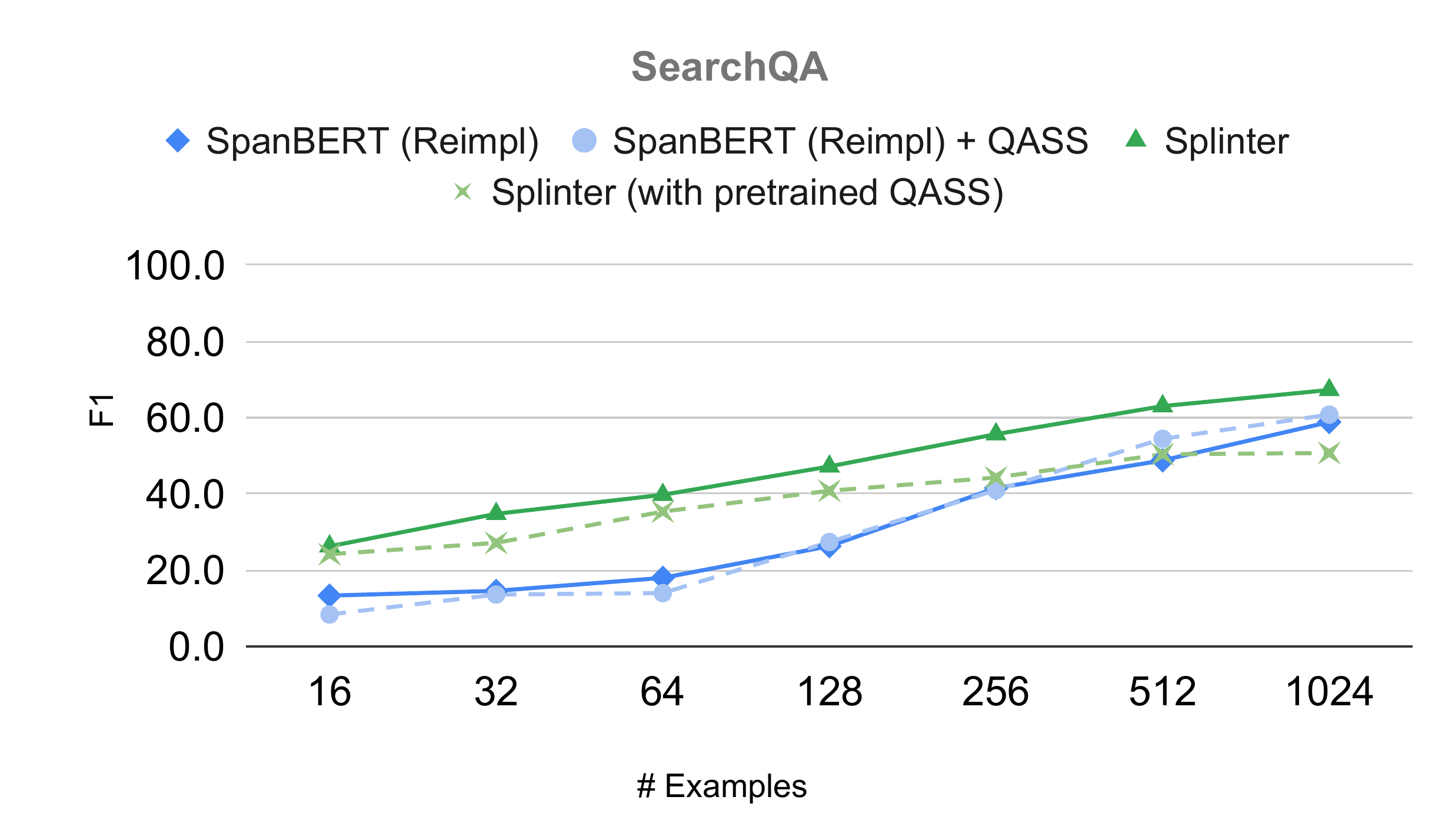}
\includegraphics[width=\columnwidth]{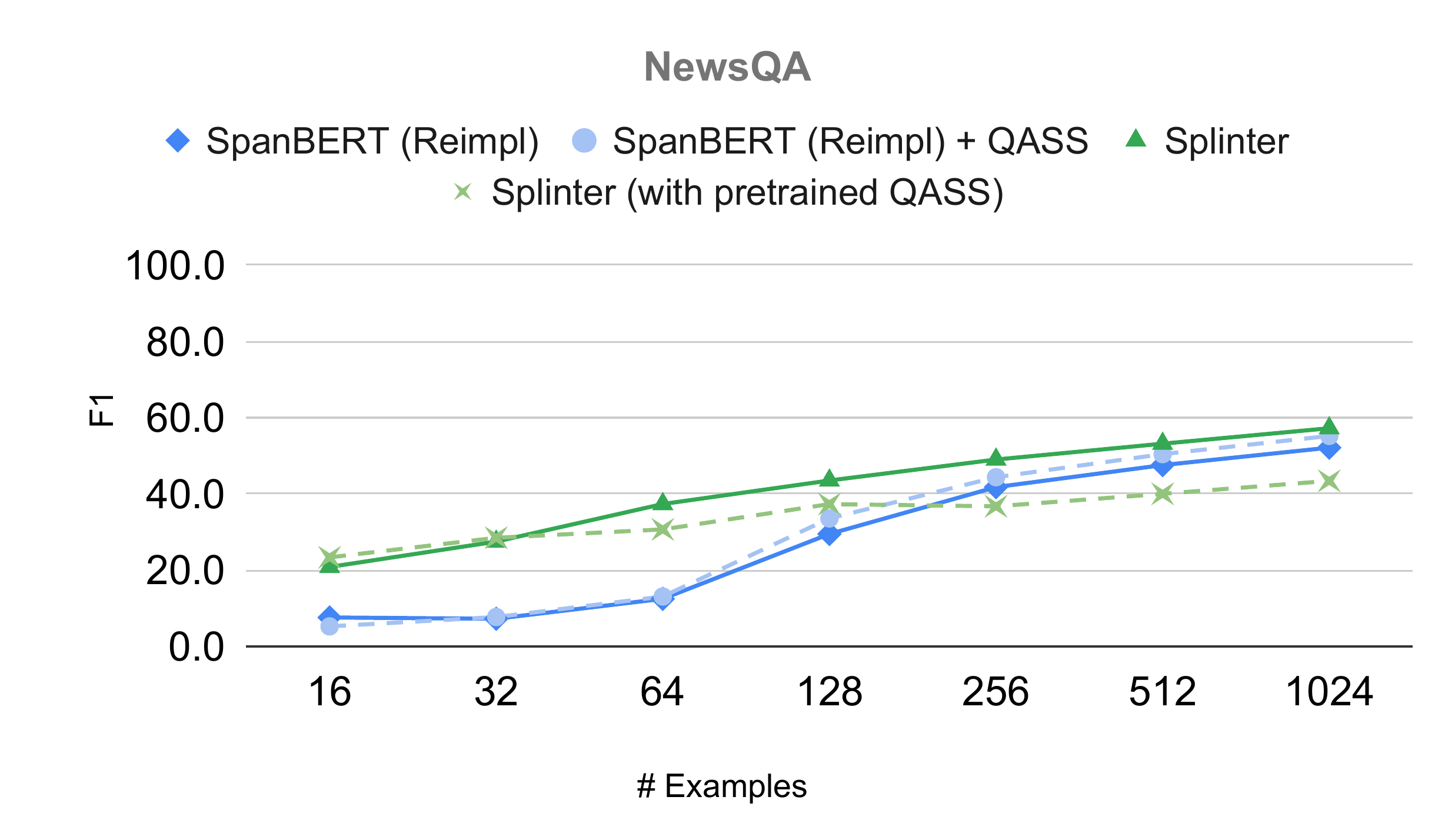}
\includegraphics[width=\columnwidth]{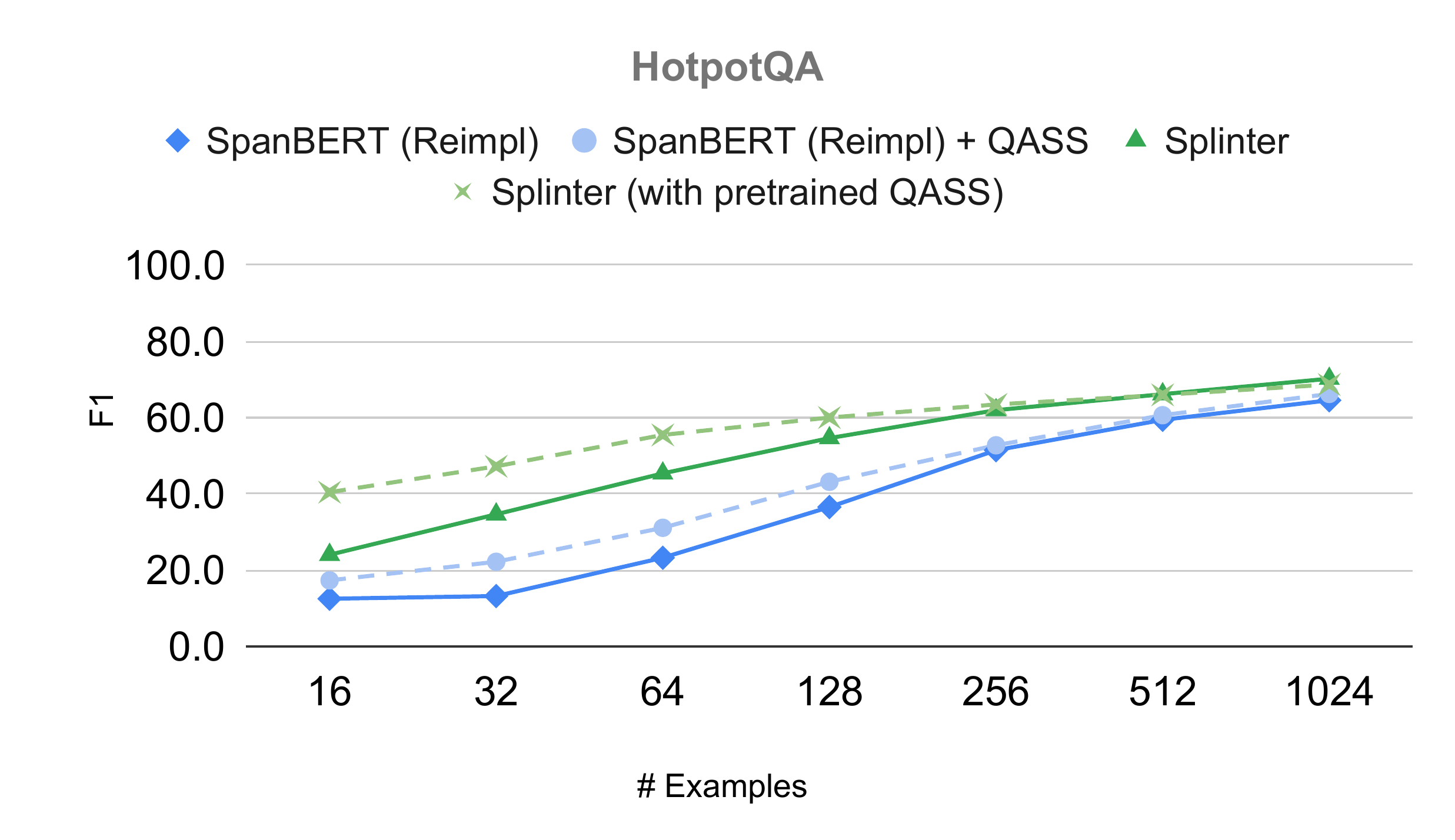}
\includegraphics[width=\columnwidth]{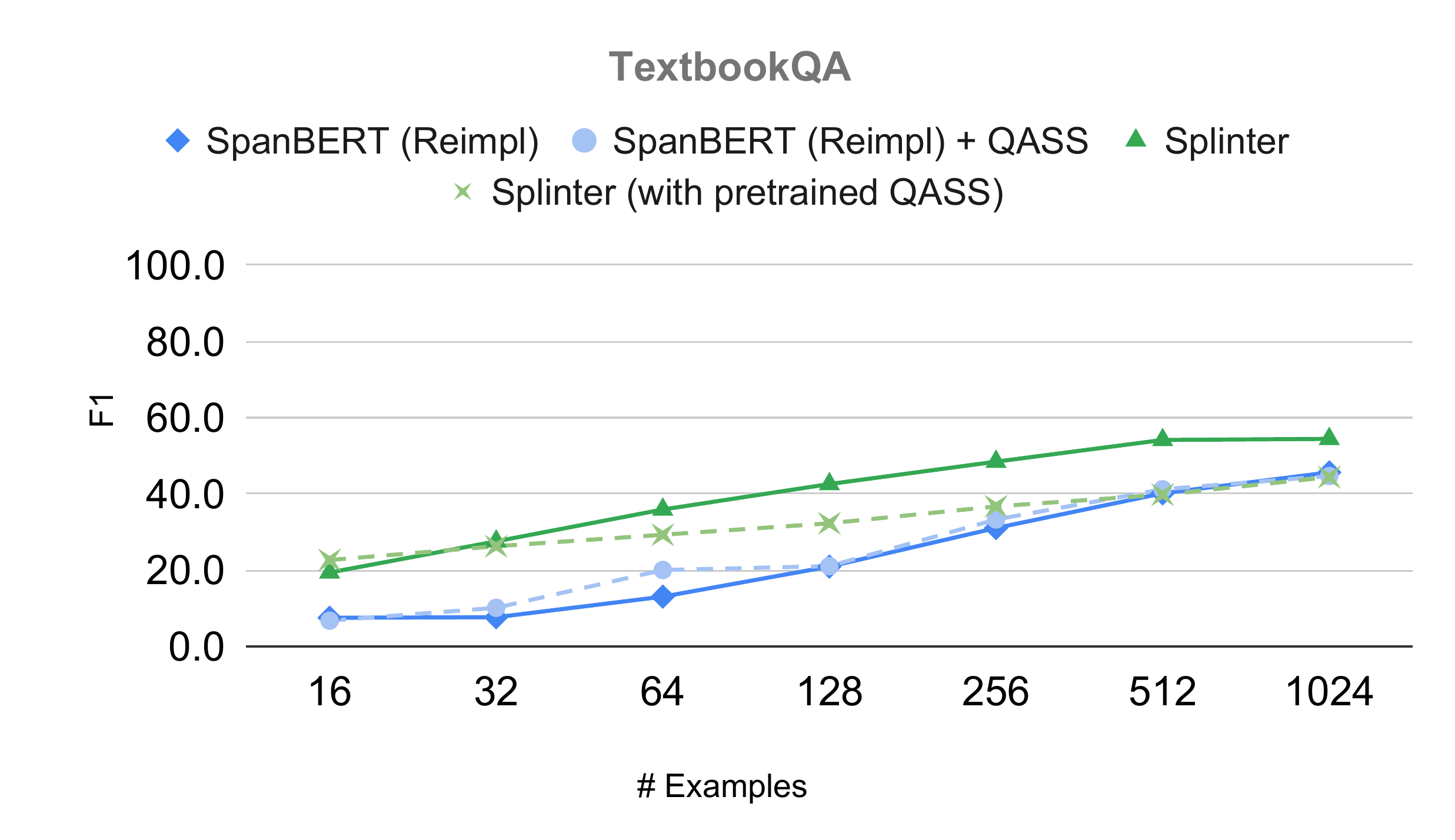}
\caption{Results complementary to ablation studies (Section 6.3). We examine the role of QASS layer by fine-tuning it on top of our SpanBERT. In addition, we test whether it is beneficial to keep the parameters of QASS from pretraining (Splinter with Head).}
\label{fig:ablation_appendix}
\end{figure*}

\paragraph{Ablation Studies} Figure~\ref{fig:ablation_appendix} shows results of ablation studies on the six question answering datasets not included in Figure~\ref{fig:ablation}.

\begin{table*}[t]
\small
\centering
\begin{tabular}{@{}lcccccccc@{}}
\toprule
\textbf{Model} & \small\textbf{SQuAD} & \small\textbf{TriviaQA} & \small\textbf{NQ} & \small\textbf{NewsQA} & \small\textbf{SearchQA} & \small\textbf{HotpotQA} & \small\textbf{BioASQ} & \small\textbf{TBQA} \\
\midrule
\small\textit{16 Examples}  & & & & & & & &  \\
\midrule
RoBERTa & ~~7.7 (4.3) & ~~7.5 (4.4) & 17.3 (3.3) & ~~1.4 (0.8) & ~~6.9 (2.7) & 10.5 (2.5) & 16.7 (7.1) & ~~3.3 (2.1) \\
SpanBERT & 12.5 (5.7) & 12.8 (5.4) & 19.7 (3.6) & ~~6.0 (1.6) & 13.0 (4.2) & 12.6 (4.3) & 22.0 (4.6) & ~~5.6 (2.5) \\
$\quad$(Reimpl) & 18.2 (6.7) & 11.6 (2.1) &19.6 (3.0) & ~~7.6 (4.1) & 13.3 (6.0) & 12.5 (5.5) & 15.9 (4.4) & ~~7.5 (2.9) \\
\textbf{Splinter} & \textbf{54.6} (6.4) & \textbf{18.9} (4.1) & \textbf{27.4} (4.6) & \textbf{20.8} (2.7) & \textbf{26.3} (3.9) & \textbf{24.0} (5.0) & \textbf{28.2} (4.9) & \textbf{19.4} (4.6) \\
\midrule
\small\textit{32 Examples}  & & & & & & & &  \\
\midrule
RoBERTa &18.2 (5.1) &10.5 (1.8) &22.9 (0.7) &~~3.2 (1.7) &13.5 (1.8) &10.4 (1.9) &23.3 (6.6) &4.3 (0.9) \\
SpanBERT &19.0 (4.6) &19.0 (4.8) &23.5 (0.9) &~~7.5 (1.3) &20.1 (3.9) &14.4 (2.9) &32.5 (3.5) &~~7.4 (1.1) \\
$\quad$(Reimpl) &25.8 (7.7) &15.1 (6.4) &25.1 (1.6) &~~7.2 (4.6) &14.6 (8.5) &13.2 (3.5) &25.1 (3.3) &~~7.6 (2.3) \\
\textbf{Splinter} & \textbf{59.2} (2.1) & \textbf{28.9} (3.1) & \textbf{33.6} (2.4) & \textbf{27.5} (3.2) & \textbf{34.8} (1.8) & \textbf{34.7} (3.9) & \textbf{36.5} (3.2) & \textbf{27.6} (4.3) \\
\midrule
\small\textit{64 Examples}  & & & & & & & &  \\
\midrule
RoBERTa &28.4 (1.7) &12.5 (1.4) &24.2 (1.0) &~~4.6 (2.8) &19.8 (2.4) &15.0 (3.9) &34.0 (1.8) &~~5.4 (1.1) \\
SpanBERT &33.6 (4.3) &22.8 (2.6) &28.4 (1.8) &~~8.8 (2.4) &26.7 (2.9) &21.8 (1.5) &43.9 (4.5) &~~7.4 (1.2) \\
$\quad$(Reimpl) &45.8 (3.3) &15.9 (6.4) &29.7 (1.5) &12.5 (4.3) &18.0 (4.6) &23.3 (1.1) &35.3 (3.1) &13.0 (6.9) \\
\textbf{Splinter} & \textbf{65.2} (1.4) & \textbf{35.5} (3.7) & \textbf{38.2} (2.3) & \textbf{37.4} (1.2) & \textbf{39.8} (3.6) & \textbf{45.4} (2.3) & \textbf{49.5} (3.6) & \textbf{35.9} (3.1) \\
\midrule
\small\textit{128 Examples}  & & & & & & & &  \\
\midrule
RoBERTa &43.0 (7.1) &19.1 (2.9) &30.1 (1.9) &16.7 (3.8) &27.8 (2.5) &27.3 (3.9) &46.1 (1.4) &~~8.2 (1.1) \\
SpanBERT &48.5 (7.3) &24.2 (2.1) &32.2 (3.2) &17.4 (3.1) &34.3 (1.1) &35.1 (4.2) &55.3 (3.8) &~~9.4 (3.0) \\
$\quad$(Reimpl) &55.8 (3.7) &26.3 (2.1) &36.0 (1.9) &29.5 (7.3) &26.3 (4.3) &36.6 (3.4) &52.2 (3.2) &20.9 (5.1) \\
\textbf{Splinter} & \textbf{72.7} (1.0) & \textbf{44.7} (3.9) & \textbf{46.3} (0.8) & \textbf{43.5} (1.3) & \textbf{47.2} (3.5) & \textbf{54.7} (1.4) & \textbf{63.2} (4.1) & \textbf{42.6} (2.5) \\
\midrule
\small\textit{256 Examples}  & & & & & & & &  \\
\midrule
RoBERTa &56.1 (5.2) &26.9 (3.5) &36.0 (3.2) &31.2 (2.4) &37.5 (1.7) &42.7 (3.1) &63.5 (1.8) &13.5 (1.9) \\
SpanBERT &55.2 (8.8) &34.0 (5.7) &41.3 (2.2) &34.7 (4.1) &42.3 (4.1) &49.4 (4.0) &67.5 (3.9) &18.2 (4.5) \\
$\quad$(Reimpl) &67.1 (2.1) &39.4 (4.0) &44.4 (3.2) &41.8 (1.8) &41.5 (3.2) &51.5 (2.8) &66.4 (2.8) &31.1 (3.4) \\
\textbf{Splinter} & \textbf{76.8} (0.6) & \textbf{57.2} (2.2) & \textbf{54.6} (1.2) & \textbf{49.0} (0.4) & \textbf{55.7} (1.9) & \textbf{62.0}  (1.6) & \textbf{77.4} (2.0) & \textbf{48.5} (2.2) \\
\midrule
\small\textit{512 Examples}  & & & & & & & &  \\
\midrule
RoBERTa &67.3 (0.7) &38.7 (3.8) &46.7 (2.2) &41.5 (2.2) &46.9 (1.6) &56.7 (1.3) &77.0 (1.9) &27.0 (2.2) \\
SpanBERT &70.0 (4.3) &44.2 (2.9) &51.5 (1.8) &42.4 (2.6) &53.9 (3.2) &61.6 (1.7) &80.3 (3.0) &33.7 (3.4) \\
$\quad$(Reimpl) &73.4 (0.4) &50.4 (2.8) &52.5 (1.9) &47.6 (1.3) &48.8 (4.1) &59.5 (1.5) &79.0 (1.9) &40.2 (0.8) \\
\textbf{Splinter} & \textbf{80.1} (0.4) & \textbf{61.9} (1.8) & \textbf{61.4} (1.1) & \textbf{53.2} (0.9) & \textbf{63.1} (1.6) & \textbf{66.2} (0.6) & \textbf{84.8} (0.9) & \textbf{54.2} (1.7) \\
\midrule
\small\textit{1024 Examples}  & & & & & & & &  \\
\midrule
RoBERTa &73.8 (0.8) &46.8 (0.9) &54.2 (1.1) &47.5 (1.1) &54.3 (1.2) &61.8 (1.3) &84.1 (1.1) &35.8 (2.0) \\
SpanBERT &77.8 (0.9) &50.3 (4.0) &57.5 (0.9) &49.3 (2.0) &60.1 (2.2) &67.4 (1.6) &89.3 (0.6) &42.3 (1.9) \\
$\quad$(Reimpl) &77.8 (0.6) &55.5 (1.9) &59.5 (1.7) &52.2 (1.2) &58.9 (1.9) &64.6 (1.2) &89.0 (1.8) &45.7 (1.5) \\
\textbf{Splinter} & \textbf{82.8} (0.8) & \textbf{64.8} (0.9) & \textbf{65.5} (0.5) & \textbf{57.3} (0.8) & \textbf{67.3} (1.3) & \textbf{70.3} (0.8) & \textbf{91.0} (1.0) & \textbf{54.5} (1.5) \\
\bottomrule
\end{tabular}
\caption{Average performance (F1) across all datasets and training set sizes. We add the standard deviation over the five seeds for each setting in parentheses. NQ and TBQA stand for Natural Questions and TextbookQA respectively. (Reimpl) stands for the SpanBERT (Reimpl) baseline (see Section 5.1).}
\label{tab:fewshot_full}
\end{table*}

\end{document}